\pdfoutput=1
\documentclass[twoside]{article}

\usepackage[accepted]{aistats2020}

\usepackage[square]{natbib}

\usepackage[usenames,dvipsnames]{xcolor}
\usepackage[utf8]{inputenc}
\usepackage{url,graphicx,subfigure,appendix,enumitem}
\usepackage{algorithm,algorithmic}
\usepackage{amsmath,amssymb,amsthm,amsfonts,latexsym,mathtools,bm,datetime,nicefrac}

\definecolor{mydarkblue}{rgb}{0,0.08,0.45}
\usepackage[colorlinks=true,
    linkcolor=mydarkblue,
    citecolor=mydarkblue,
    filecolor=mydarkblue,
    urlcolor=mydarkblue]{hyperref} 
\usepackage[capitalize]{cleveref}

\usepackage[usenames,dvipsnames]{xcolor}
\usepackage[disable]{todonotes}

\newcommand{\todob}[2][]{\todo[color=Red!20,size=\tiny,inline,#1]{B: #2}} 
 
\newcommand{\abas}[1]{\todo[color=Green!20,size=\tiny,inline]{Abbas: #1}}

\usepackage[compact]{titlesec}
\titlespacing{\section}{10pt}{*0}{*0}
\titlespacing{\subsection}{12pt}{*0}{*0}

\usepackage{setspace}
\newcommand{\muhat}{\widehat{\mu}}
\newcommand{\eps}{\varepsilon}

\newcommand{\cA}{\mathcal{A}}

\newcommand{\cH}{\mathcal{H}}
\newcommand{\cL}{\mathcal{L}}

\newtheorem{theorem}{Theorem}
\newtheorem{lemma}{Lemma}
\newtheorem{proposition}{Proposition}

\DeclareMathOperator*{\argmax}{arg\,max}
\DeclareMathOperator*{\argmin}{arg\,min}

\newcommand{\hth}{{\hat \theta_t}}

\newcommand{\sth}{\theta^\star}
\newcommand{\thetastar}{\sth}

\newcommand{\norm}[1]{\left\|#1\right\|}

\newcommand{\indnorm}[2]{\left\|#1\right\|_{#2}}

\newcommand{\invMt}{M_{t}^{-1}}
\newcommand{\invHt}{H_{t}^{-1}}

\newcommand{\ind}[1]{\mathbf{I}\left\{#1\right\}}
\newcommand{\E}{\mathbf{E}}

\newcommand{\Ex}[1]{\E\left[#1\right]}
\renewcommand{\Pr}{\mathbf{P}}
\renewcommand{\P}{\Pr}
\newcommand{\pr}[1]{\P\left(#1\right)}
\newcommand{\Prob}[1]{\P\left(#1\right)}

\newcommand{\R}{\mathbb{R}}
\newcommand{\C}{\mathcal{C}}
\newcommand{\transpose}{^{\mathsf{T}}}
\newcommand{\dop}[2]{\left\langle #1, #2 \right\rangle}
\newcommand{\thetahat}{\widehat{\theta}}
\newcommand{\ftilde}{\widetilde{f}}

\renewcommand{\epsilon}{\varepsilon}
\renewcommand{\tilde}{\widetilde}
\renewcommand{\hat}{\widehat}

\newcommand{\randucb}{{\tt RandUCB}}

\newcommand{\notetoself}[1]{}

\newcommand{\ls}{\text{ls}}
\newcommand{\conc}{\text{conc}}
\newcommand{\anti}{\text{anti}}
\newcommand{\mle}{\text{mle}}
\newcommand{\bound}{\text{bound}}
\begin{document}

\runningtitle{Randomized UCB for Bandit Problems}
\runningauthor{Vaswani, Mehrabian, Durand, Kveton}

\twocolumn[
\aistatstitle{Old Dog Learns New Tricks: Randomized UCB for Bandit Problems}
\aistatsauthor{Sharan Vaswani \And Abbas Mehrabian \And  Audrey Durand \And Branislav Kveton}
\aistatsaddress{Mila, Universit\'e de Montr\'eal
\And McGill University 
\And Mila, Universit\'e Laval\\
\And Google Research
} 
]

\begin{abstract}
We propose $\randucb$, a bandit strategy that builds on theoretically derived confidence intervals similar to upper confidence bound (UCB) algorithms, but akin to Thompson sampling (TS), it uses randomization to trade off exploration and exploitation. In the $K$-armed bandit setting, we show that there are infinitely many variants of $\randucb$, all of which achieve the minimax-optimal $\widetilde{O}(\sqrt{K T})$ regret after $T$ rounds. Moreover, for a specific multi-armed bandit setting, we show that both UCB and TS can be recovered as special cases of $\randucb$. For structured bandits, where each arm is associated with a $d$-dimensional feature vector and rewards are distributed according to a linear or generalized linear model, we prove that $\randucb$ achieves the minimax-optimal $\widetilde{O}(d \sqrt{T})$ regret even in the case of infinitely many arms. Through experiments in both the multi-armed and structured bandit settings, we demonstrate that $\randucb$ matches or outperforms TS and other randomized exploration strategies. Our theoretical and empirical results together imply that $\randucb$ achieves the best of both worlds. 
\end{abstract}
\section{Introduction}
\label{sec:introduction}
The \emph{multi-armed bandit} (MAB)~\citep{woodroofe1979one,lai1985asymptotically,auer2002finite} is a sequential decision-making problem with \emph{arms} corresponding to actions available to a \emph{learning agent} to choose from. For example, the arms may correspond to potential treatments in a clinical trial or ads available for display on a website. When an arm is chosen (\emph{pulled}), the agent receives a \emph{reward} from the \emph{environment}. In the stochastic MAB, which is our focus, this reward is sampled from an underlying distribution associated with that particular arm. The agent's goal is to maximize its expected reward accumulated across interactions with the environment (\emph{rounds}). As the agent does not know the arms' reward distributions, she faces an \emph{exploration-exploitation dilemma}: \emph{explore} and learn more about the arms, or \emph{exploit} and choose the arm with the highest estimated mean thus far. 

\emph{Structured bandits}~\citep{li2010contextual,filippi2010parametric, abbasi2011improved,agrawal2013thompson, li2017provable} are generalizations of the MAB problem in which each arm is associated with a known \emph{feature} vector. These features encode properties of the arms; for example, they may represent the properties of a drug being tested in a clinical trial, or the meta-data of an advertisement on a website. In structured bandits, the expected reward of an arm is an unknown function of its feature vector. 

This function is often assumed to be parametric; an important special case is the \emph{linear bandit}~\citep{dani2008stochastic,rusmevichientong2010linearly,abbasi2011improved}, where the function is linear and the expected reward is the dot product of the feature vector and an unknown parameter vector. Similarly, in the \emph{generalized linear bandit}~\citep{filippi2010parametric,li2017provable,kveton2019randomized}, the expected reward follows a generalized linear model~\citep{mccullagh1984generalized}. 

\subsection{Classic exploration strategies}
In both the multi-armed and structured bandit settings, classic strategies to trade off exploration and exploitation include \emph{$\varepsilon$-greedy} (EG)~\citep{sutton1998reinforcement,auer2002finite}, \emph{optimism in the face of uncertainty} (OFU)~\citep{auer2002finite,abbasi2011improved},  and \emph{Thompson sampling} (TS)~\citep{thompson1933likelihood,agrawal2013further}.
The EG policy is simple, can be applied to any MAB or structured bandit setting, and  is thus widely used in practice. However, it is statistically sub-optimal, does not explore in a problem dependent manner , and its practical performance is sensitive to hyper-parameter tuning. 
On the other hand, deterministic strategies based on OFU, such as the celebrated UCB1 algorithm~\citep{auer2002finite}, construct closed-form high-probability confidence sets. OFU-based algorithms are theoretically optimal in many bandit settings, including MAB and linear bandits. However, since these confidence sets are constructed to obtain good worst-case performance, they often have poor empirical performance on typical problem instances. Moreover, for structured bandits, when the feature-reward mapping is non-linear (e.g., generalized linear models), we can only construct coarse confidence sets~\citep{filippi2010parametric,zhang2016online,jun2017scalable,li2017provable}, which are often too conservative in practice.

In contrast, TS is a randomized strategy that  maintains a posterior distribution over the unknown parameters, and samples from it in order to choose actions. When the posterior has a closed form, as in the Bernoulli or Gaussian MAB or linear bandits, it is possible to sample exactly from it. In these cases, TS is computationally efficient and have good empirical performance~\citep{chapelle2011empirical}. However, when there is no closed form posterior, one has to resort to approximate sampling techniques, which are typically expensive~\citep{gopalan2014thompson,kawale2015efficient,riquelme2018deep} and limit the practical applicability of TS. From a theoretical point of view, TS results in near-optimal regret bounds for the MAB problem~\citep{agrawal2013further}, but current analyses result in a sub-optimal dependence on the feature dimension for structured bandits~\citep{ abeille2017linear,agrawal2013thompson}. 

\subsection{Randomized exploration strategies}
There has been substantial recent research on using \emph{bootstrapping}~\citep{baransi2014sub,eckles2014thompson,osband2015bootstrapped,tang2015personalized,elmachtoub2017practical,vaswani2018new} or designing general randomized exploration schemes~\citep{kveton2019perturbedlinear,kveton2019perturbedmab, kveton2019garbage, kveton2019randomized, kim2019optimality}. These data-driven strategies do not rely on problem-specific confidence sets, neither do they require a posterior distribution. 
Moreover, they are applicable even when the feature to reward mappings is a complex one (e.g., a neural networks)~\citep{osband2015bootstrapped,vaswani2018new,kveton2019garbage}. 

However, these strategies suffer from theoretical and practical drawbacks. In particular, for typical bootstrapping strategies, theoretical guarantees have been derived only for linear bandits and MAB with Gaussian or Bernoulli rewards~\citep{lu2017ensemble,osband2015bootstrapped, vaswani2018new}. General randomized strategies~\citep{ kveton2019perturbedmab,kveton2019garbage, kim2019optimality} achieve near-optimal regret bounds in the general MAB setting; however, the degree of exploration is difficult to control, complicating their proofs. For structured bandits, randomized strategies have been proposed in the linear~\citep{kveton2019perturbedlinear} and generalized linear~\citep{kveton2019randomized}  settings. However, their analysis for linear bandits closely follows that of TS and inherits its sub-optimality in the feature dimension~\citep{kveton2019perturbedlinear}, and proving regret bounds for the generalized linear case~\citep{kveton2019randomized} requires additional assumptions.

From a practical perspective, the advantage of these randomized strategies is that they do not rely on closed form posterior distributions like TS, but they ``sample'' from an implicit distribution. This distribution could be induced via bootstrapping~\citep{osband2015bootstrapped, lu2017ensemble, vaswani2018new}, adding pseudo-observations~\citep{kveton2019garbage}, or randomizing the observed data~\citep{kveton2019perturbedmab,kveton2019perturbedlinear,kveton2019randomized,kim2019optimality}. These choices complicate the resulting algorithm. Moreover, in order to generate a ``sample'', these strategies require solving a maximum likelihood estimation problem in each round. Unlike computing an upper confidence bound (as in OFU) or sampling from the posterior (as for TS), this estimation problem cannot be solved in an efficient, online manner while preserving regret guarantees~\citep{jun2017scalable}. For computational efficiency, these strategies resort to heuristics for approximating the maximum likelihood estimator (MLE)~\citep{vaswani2018new, kveton2019garbage,osband2015bootstrapped, lu2017ensemble}. Unfortunately, these approximations do not have rigorous theoretical guarantees and add another layer of complexity to the algorithm design.
\section{Our Contribution}
\label{sec:contributions}
As general randomized strategies are complicated and computationally expensive even in the standard MAB or structured bandit settings, we consider randomizing simple OFU-based algorithms. To this end, we propose the $\randucb$ (meta-)algorithm, which relies on existing theoretically derived confidence sets, but similar to TS, it uses randomization to trade off exploration and exploitation. 
In \cref{sec:general-algorithm}, we describe the general framework of the $\randucb$ meta-algorithm.

In \cref{sec:MAB}, we  instantiate $\randucb$ in the MAB setting. We show that TS can be viewed as a special case of $\randucb$ in a specific MAB setting (\cref{sec:MAB-connection}). Furthermore, by reasoning about the algorithmic choices in $\randucb$, we derive variants of classic exploration strategies. For example, we formulate \emph{optimistic Thompson sampling}, a variant of TS which only generates posterior samples greater than the mean, and show that it results in comparable theoretical and empirical performance as TS (\cref{app:expes:ots}). More generally, we show that there are infinitely many variants of $\randucb$, all of which achieve the minimax-optimal $\widetilde{O}(\sqrt{K T})$ regret for an MAB  with $K$ arms over $T$ rounds (\cref{sec:MAB-theory}).

For structured bandits, we present an instantiation of the $\randucb$ meta-algorithm when the rewards follow a linear (\cref{sec:LB}) or a generalized linear model (\cref{sec:GLB}). We show that $\randucb$ achieves the optimal $\widetilde{O}(d \sqrt{T})$ regret for $d$-dimensional feature vectors, even with infinitely many arms. In both these settings, $\randucb$ matches the theoretical regret bounds of the corresponding OFU-based algorithms~\citep{abbasi2011improved, li2017provable} up to constant factors. To the best of our knowledge, $\randucb$ is the first randomized algorithm that results in the near-optimal dependence on the dimension in the infinite-armed case. For all the above settings, the algorithm design of $\randucb$ enables simple proofs that extend naturally from the existing TS and OFU analyses. 

Finally, we conduct experiments in the MAB and structured bandit settings\footnote{See code: \url{https://github.com/vaswanis/randucb}.}, investigating the impact of algorithmic design choices through an ablation study (\cref{app:expes:ablation}), and demonstrating the practical effectiveness and efficiency of $\randucb$ (\cref{sec:experiments}). In all settings, the performance of $\randucb$ is either comparable to or better than that of TS and the more complex, computationally expensive generalized randomized strategies. 
\section{The $\randucb$ Meta-Algorithm}
\label{sec:general-algorithm}
In this section, we describe the general form of $\randucb$ and detail the design decisions. 
Consider a bandit setting with action set $\cA$. When arm $i \in \cA$ is pulled, a reward is drawn from its underlying distribution, 
with  mean $\mu_i$ and support $[0, 1]$, and is presented to the learner. 
The learner's objective is to maximize its expected cumulative reward across $T$ rounds. 

An OFU-based bandit algorithm keeps track of the estimated mean $\muhat_i(t)$, defined as the average of rewards received from arm $i$ until round $t$. The algorithm also maintains a confidence interval of size $\C_i(t)$ around the estimated mean.
The value of $\C_i(t)$ decreases as an arm is pulled more, and indicates how accurate $\muhat_i(t)$ is at estimating $\mu_i$.

Although the exact values of $\muhat_i(t)$ and $\C_i(t)$ depend on the bandit setting under consideration, OFU-based strategies~\citep{auer2002finite, abbasi2011improved} have the same general form: in round $t$, they choose the arm
\begin{align}
\label{eqn:generic:ucb}
i_t & = \argmax_{i \in \cA} \left\{ \muhat_i(t) + \beta \; \C_i(t) \right\}. 
\end{align}
The parameter $\beta$ is carefully chosen to trade off exploration and exploitation optimally. We will instantiate this algorithm for the multi-armed (\cref{sec:MAB}), linear (\cref{sec:LB}), and generalized linear (\cref{sec:GLB}) bandit settings. As a simple modification, $\randucb$ randomizes the confidence intervals and chooses the arm
\begin{align}
\label{eqn:generic:randucb}
i_t & = \argmax_{i\in\cA} \left\{ \muhat_i(t)  + Z_t \; \C_i(t) \right\},
\end{align}
where the deterministic quantity $\beta$ is replaced by a random variable $Z_t$. Here, $Z_1,\dots, Z_T$ are i.i.d.\ samples from the \emph{sampling distribution} that we describe next.  

\subsection{The sampling distribution}
\label{sec:randucb:sampling_dist}
The random variables $Z_1,\dots,Z_T$ are i.i.d.\ and have the same distribution as a template random variable $Z$, explained below. We consider a discrete distribution for $Z$ on the interval $[L, U]$, supported on $M$ points. Let $\alpha_1=L,\dots,\alpha_M = U$ denote $M$ equally spaced points in $[L, U]$, and define $p_m \coloneqq \Prob{Z=\alpha_m}$. If $M = 1$ and $L = U = \beta$, then we recover the OFU-based algorithm, Eq.~\eqref{eqn:generic:ucb}. If $L = 0$ and $U = \beta$, then $\randucb$ chooses between values in the $[0, \beta]$ range; in this case, the $\alpha_m$ can be viewed as \emph{nested confidence intervals}. 

We choose a constant value for $M$ throughout this paper, but note that letting $M \to \infty$ can simulate a fine discretization of an underlying continuous distribution supported on $[L,U]$. To obtain optimal theoretical guarantees, the probabilities $p_1, \ldots, p_M$ in $\randucb$ must be chosen in a way that ensures $\Prob{Z \geq \beta} > 0$. This guarantees that the algorithm has enough optimism and we will later prove that this constraint ensures that $\randucb$ attains optimal regret for all the bandit settings we consider. 

Our choice of the sampling distribution (the $p_m$ values) is inspired from a Gaussian distribution truncated in the $[0,U]$ interval and has tunable hyper-parameters $\epsilon,\sigma>0$. The former is the constant probability to be put on the highest point: $\alpha_M=U$ with $p_M = \epsilon$. For the remaining $M - 1$ points, we use a discretized Gaussian distribution; formally, for $1\leq m \leq M-1$, let $\overline{p_m} \coloneqq \exp(-\alpha_m^2/2\sigma^2)$ and let $p_m$ denote the normalized probabilities, that is, $p_m\coloneqq (1-\epsilon) \; \overline{p_m}/(\sum_m \overline{p_m})$. The above choice can be viewed as a truncated (between $0$ and $U$) and discretized (into $M$ points) Gaussian distribution. As we explain in \cref{sec:MAB-connection}, choosing this distribution resembles Gaussian TS.\footnote{One might also consider a discretized uniform distribution on $[0,U]$, but our experiments in~\cref{app:expes} show that this choice performs poorly in practice.}

\subsection{Algorithmic decisions}
\label{sec:randucb:alg_decisions}
\paragraph{Optimism} By only considering positive values for $Z$ (by setting $L = 0$), we maintain the OFU principle~\citep{auer2002finite, abbasi2011improved} of the corresponding OFU-based algorithm. Although our theoretical results allows $Z$ to take negative values, we experimentally observe that this does not significantly improve the empirical performance of $\randucb$~(see Figure~\ref{fig:results:ablation:optimism_coupling}
in~\cref{app:expes:ablation}).

\paragraph{Coupling the arms} By default, in each round $t$, $\randucb$ samples a single value of $Z_t$ that is shared between all the arms (see Eq.~\eqref{eqn:generic:randucb})  thus ``coupling'' the arms. Alternatively, we could consider \emph{uncoupled $\randucb$} where in each round $t$, each arm $i$ generates its own independent copy of $Z$, say $Z_{i,t}$, and the algorithm selects the arm $i_t  = \argmax_i \left\{ \muhat_i(t)  + Z_{i,t} \; \C_i(t) \right\}$. This is similar to the Boltzmann exploration algorithm in~\citet{cesa2017boltzmann}. However, our experiments show that the uncoupled variant does not perform better than the default, coupled version~(see Figure~\ref{fig:results:ablation:optimism_coupling}
in~\cref{app:expes:ablation}).

In the next sections, we revisit these decisions, instantiate $\randucb$, and analyze its performance in specific bandit settings.
The subsequent theoretical results hold for $L=0$,
any positive integer $M$,
and any positive constants $\epsilon$ and $\sigma$.
The value of $U$ depends on the specific bandit setting.
For the empirical evaluation (\cref{sec:experiments} and~\cref{app:expes}), the specific values of $L$, $U$, $M$, $\epsilon$, and $\sigma$ will be specified for each experiment.
\section{Multi-Armed Bandit}
\label{sec:MAB}
In this section, we consider a stochastic multi-armed bandit (MAB) with $\vert \cA \vert = K$ arms. Without loss of generality, we may assume that arm $1$ is optimal, namely $\mu_1 = \max_{i} \mu_i$, and refer to $\Delta_i = \mu_1 - \mu_i$ as the \emph{gap} of arm $i$. Maximizing the expected reward is equivalent to minimizing the \emph{expected regret} across $T$ rounds. If a bandit algorithm pulls arm $i_t$ in round $t$, then it incurs an expected (cumulative) regret of
\(R(T) \coloneqq 
\sum_{t=1}^{T}
\E[\mu_1 - \mu_{i_t}]
=
\sum_{t=1}^{T}
\E[\Delta_{i_t}].
\)
\subsection{Instantiating \randucb}
\label{sec:MAB-algorithm}
Let $s_i(t)$ denote the number of pulls and $Y_i(t)$ denote the total reward received from arm $i$ by round $t$. Then the estimated mean is simply $\muhat_i(t)=Y_i(t) / s_i(t)$ (we set $\muhat_i(t)=0$  if arm $i$ has never been pulled). The confidence interval corresponds to the standard deviation in the estimation of $\mu_i$ and is given as $\C_i(t) = \sqrt{\frac{1}{s_i(t)}}$. To ensure that $s_i(t)>0$, $\randucb$ begins by pulling each arm once and in each subsequent round $t > K$, selects
\begin{align}
i_t = \argmax_i \left\{ \muhat_i(t)  + Z_t \; \sqrt{\frac{1}{s_i(t)}} \right\}.
\label{eqn:randucb-mab}
\end{align}
The corresponding OFU-based algorithm~\citep[Figure~1]{auer2002finite} sets the constant $\beta = \sqrt{2 \ln(T)}$. For $\randucb$, we choose $L = 0$ and $U = 2 \sqrt{\ln(T)}$, that is, we inflate\footnote{This inflation is a technicality needed for our analysis.} the confidence interval by $\sqrt{2}$.

\subsection{Connections to TS and EG}
\label{sec:MAB-connection}
We now describe how $\randucb$ relates to existing algorithms.
Recall that TS~\citep{agrawal2013further} may draw samples below the empirical mean for each arm, whereas $\randucb$ with $L \geq 0$ samples from a one-sided distribution above the mean. In order to make the connection from $\randucb$ to TS, we consider a variant of TS which only samples values above the mean for each arm,\footnote{It samples from a conditional posterior distribution, conditioned on the sample being larger than the mean.} referred to as \emph{optimistic Thompson sampling} (OTS). Our experiments show that OTS has similar empirical performance as TS~(\cref{app:expes:ots}). We show that $\randucb$ with $M\to\infty$ approaches OTS with a Gaussian prior and posterior. First, observe that uncoupled $\randucb$ with $Z \sim \mathcal{N}(0,1)$ without truncation or discretization exactly corresponds to TS. Now consider optimistic TS and further truncate the tail of the Gaussian posterior at $2 \sqrt{\ln(T)}$. By putting a constant probability mass of $\epsilon$ at $2\sqrt{\ln T}$ (the upper bound of the distribution) and discretizing the resulting distribution at $M - 1$ equally-spaced points, we obtain the {\em uncoupled} variant of $\randucb$.

The flexibility of $\randucb$ also allows us to consider a variant that resembles an adaptive $\epsilon$-greedy strategy. Recall that the classical $\epsilon$-greedy (EG) strategy~\citep{auer2002finite, langford2008epoch} chooses a random action with probability $\epsilon$ and the greedy action with probability $1 - \epsilon$. For a constant $\epsilon$, EG might result in linear regret, whereas decreasing $\epsilon$ over time results in a sub-optimal $O(T^{2/3})$ regret~\citep{auer2002finite}. An \emph{adaptive $\epsilon$-greedy} can be instantiated from $\randucb$ as follows: 
let $Z$ be a random variable that takes value 0 with probability $1-\epsilon$ and $2\sqrt{\ln T}$ with probability $\epsilon$.

This results in choosing the greedy action with probability $1 - \epsilon$ and choosing the action that maximizes the data-dependent upper-confidence-bound with probability $\epsilon$. 
Theorem~\ref{thm:reg-randucb-mab} below implies that the regret of this modification of the $\eps$-greedy algorithm is bounded by $O(\sqrt{KT\ln(KT)})$.

\subsection{Regret of $\randucb$ for MAB}
\label{sec:MAB-theory}
In this section, we first bound the regret of the default optimistic, coupled variant of $\randucb$ with a general distribution for $Z$ and then obtain a bound for the uncoupled variant.

\begin{theorem}[Minimax regret of $\randucb$ with coupled arms for MAB]
\label{thm:reg-randucb-mab}
Let $c_1 \coloneqq 1+\sqrt{\ln(KT^2)}$ and
$c_3  \coloneqq 2K \ln\left(1 + \frac{T}{K}\right)$.
For any $c_2>c_1$, the regret $R(T)$ of $\randucb$ for MAB is bounded by
\begin{align*}
& (c_1 + c_2) \left(1 + \frac{2}{\Prob{Z>c_1} - \Prob{|Z|> c_2}} \right) \times \sqrt{c_3 T} \\ & + T \;  \Prob{|Z|> c_2} +K+1.
\end{align*}
\end{theorem}
The proof for the above theorem uses a reduction from linear bandits; we defer it to \cref{sec:linear:mab_proof}.

The above result implies that the regret of $\randucb$ can be bounded by $O(\sqrt{KT\ln(KT)})$ so long as \textbf{(i)} $\pr{Z > 1+\sqrt{\ln(KT^2)}}>0$ and \textbf{(ii)} $|Z| \leq c_2$ deterministically. 
By choosing $U=2\sqrt{\ln T}$,
our sampling distribution would lie in $[0, 2 \sqrt{\ln(T)}]$, so condition (ii) holds by setting $c_2 =  2 \sqrt{\ln T}$ in Theorem~\ref{thm:reg-randucb-mab}. Since the considered sampling distribution in~\cref{sec:randucb:sampling_dist} has a constant probability mass of $\epsilon$ at $U=2 \sqrt{\ln(T)}$ by design, it ensures that $\Prob{Z > c_1}$ is a positive constant.  Since any consistent algorithm for MAB has regret at least $\Omega(\sqrt{KT})$ (see, e.g., \citet[Theorem~15.2]{lattimore2018bandit}), $\randucb$ is minimax-optimal up to logarithmic factors.

The next result states that  uncoupled $\randucb$ achieves problem-dependent logarithmic regret, therefore also being nearly-optimal.
\begin{theorem}[Instance-dependent regret of uncoupled $\randucb$ for MAB]
\label{thm:instancedependent}
If $Z$ takes $M$ different values $0\leq\alpha_1\leq\cdots\leq\alpha_M$ with probabilities $p_1, p_2, \ldots, p_M$, the regret $R(T)$ of uncoupled $\randucb$ can be bounded as
\(
O\left(\sum_{\Delta_i>0} \Delta_i^{-1}\right)\times\left(\frac{M}{p_M}+ T e^{-2\alpha_M^2} + \alpha_M^2\right).
\)
\end{theorem}
Since the sampling distribution of $\randucb$ satisfies $U=\alpha_M=2\sqrt{\ln T}$ and $M$ and $p_M$ are constant, uncoupled $\randucb$ attains the optimal instance-dependent regret $O\left(\ln T \times \left(\sum \Delta_i^{-1}\right)\right)$ (see, e.g., \citet[Theorem~16.4]{lattimore2018bandit}). By a standard reduction, Theorem~\ref{thm:instancedependent} implies that uncoupled $\randucb$ achieves the problem-independent $\tilde{O}(\sqrt{KT})$ regret. Please
refer to~\cref{sec:instancedependentupperbound} for the proof and the statement of~\cref{thm:rucbmainregbound} for a tighter regret bound. 
\section{Structured Bandits}
\label{sec:structured-bandits}
In this section, we consider the structured bandit setting where each arm is  associated with a $d$-dimensional feature vector and there exists an underlying parametric function that maps these features to rewards. Let $x_i\in\R^d$ denote the corresponding feature vector for arm $i\in \cA$. We assume that $d>1$ and $\|x_i\|\leq1$ for every arm $i$. We also assume that the function mapping a feature vector to the expected reward is parameterized by an unknown parameter vector $\theta^\star$ with $\|\theta^\star\|\leq1$, and that the rewards lie in $[0,1]$.\footnote{It is easy to generalize our results to reward distributions bounded in any known interval. Similarly, the analyses can be adapted to handle sub-gaussian distributions.} We first consider the linear feature-reward mapping. 

\subsection{Linear bandits}
\label{sec:LB}
In linear bandits, the expected reward of an arm is the dot product of its corresponding feature vector and the unknown parameter. Formally, if $Y_t$ is the reward obtained in round $t$, then $\E[Y_t | i_t = i] = \dop{x_i}{\theta^\star}$. If $i_t$ is the arm pulled in round $t$ and arm $1$ is the optimal arm, then the regret can be defined similarly as in the MAB case, but with an ``effective'' gap $\Delta_i = \dop{x_1 - x_i}{\theta^\star}$,
\begin{align}\label{eqn:lb:regret}
R(T) := \sum_{t = 1}^T  \E\left[ \dop{x_1-x_{i_t}}{\sth}  \right]=\sum_{t = 1}^T  \E [\Delta_{i_t}]. \end{align}
Let us denote $X_t\coloneqq x_{i_t}$ and define the Gram matrix $M_t \coloneqq \lambda I_d + \sum_{\ell=1}^{t-1} X_{\ell}X_{\ell}\transpose$. Here, $\lambda>0$ is the {\em $\ell_{2}$ regularization parameter}. We define the norm $\|x\|_M \coloneqq \sqrt{x\transpose M x}$ for any positive definite $M$. 

\subsubsection{Instantiating \randucb}
\label{sec:LB-algorithm}
Given the observations $(X_\ell, Y_\ell)_{\ell = 1}^{t-1}$ gathered until round $t$, the maximum likelihood estimator (MLE) for linear regression is $\thetahat_t \coloneqq M_t^{-1}\sum_{\ell=1}^{t-1} Y_{\ell}X_{\ell}$. The estimated mean for the reward of arm $i$  is $\muhat_i(t) = \langle \hth, x_i \rangle$ and the corresponding confidence interval is $\C_i(t) = \indnorm{x_i}{\invMt}$. Thus, $\randucb$ chooses arm 
\begin{align}
i_t\coloneqq \argmax_{i \in \cA} \left\{ \langle \hth, x_i \rangle + Z_t \indnorm{x_i}{\invMt}
\right\}.
\label{eq:randucb-lb}
\end{align}
Note that the corresponding OFU-based algorithm~\citep[Theorem~2]{abbasi2011improved} sets $\beta = \sqrt{\lambda}+\frac12\sqrt{\ln (T^2\lambda^{-d} \det(M_t) )}$. We prove the following theorem for $\randucb$. 

\subsubsection{Regret of $\randucb$ for linear bandits}
\label{sec:theoreticalresultsforlinearbandits}
\begin{theorem}
\label{thm:theoreticalresultsforlinearbandits}
Let $c_1 = \sqrt{\lambda}+\frac12\sqrt{d\ln\left(T+T^2/d\lambda\right)}$ and $c_3  \coloneqq 2d \ln\left(1 + \frac{T}{d \lambda}\right)$.
For any $c_2>c_1$, the regret of $\randucb$  for linear bandits is bounded by
\begin{align*}
& (c_1 + c_2) \left(1 + \frac{2}{\Prob{Z>c_1} - \Prob{|Z|> c_2}} \right) \times \sqrt{c_3 T} \\& + T \; \Prob{|Z|> c_2} +1.
\end{align*}
\end{theorem}
\begin{proof}
Let $\ftilde_t(x) \coloneqq \langle \hth, x \rangle + Z_t \indnorm{x}{\invMt}$
and define the events
\begin{align*}
& E^{\ls} \coloneqq \left\{ \forall i \in [K],\forall t \in [T]
\text{;} \quad \vert \langle x_i, \hth - \sth \rangle
\vert \leq c_1 \indnorm{x_i}{\invMt} \right\}, \\
& E^{\conc}_{t} \coloneqq \left\{ \forall i \in [K] \text{;} \quad \vert \tilde{f}_{t}(x_i) - \langle x_i, \hth  \rangle \vert \leq c_2 \indnorm{x_i}{\invMt} \right\}, \\
& E^{\anti}_{t} \coloneqq \left\{ \tilde{f}_{t}(x_1) - \langle x_1, \hth  \rangle > c_1 \indnorm{x_1}{\invMt} \right\},
\end{align*}
and assume for now that we have the following bounds for their probabilities: $\Prob{E^{\ls}} \geq 1-p_1$, $\Pr(E^{\conc}_t) \geq 1-p_2$, and $\Pr(E^{\anti}_t) \geq p_3$.

In \cref{sec_reg_linear}, we prove an upper bound for the regret of any {\em index-based algorithm} in terms of $p_1$, $p_2$, and $p_3$. 
An index-based algorithm is one that in each round $t$  chooses the arm $i_t$ that maximizes some function $\ftilde_t(x)$, i.e., $i_t = \argmax_{i} \ftilde_t(x_i)$. Theorem~\ref{thm:linbanditsmainthm} in \cref{sec_reg_linear} bounds the regret of such an algorithm by
\begin{align}
(c_1 + c_2) \left(1 + \frac{2}{p_3 - p_2} \right) \sqrt{c_3 T} + T \; (p_1 + p_2).
\label{genericregretbound}
\end{align}
For \randucb, we have $\ftilde_t(x) = \langle \hth, x \rangle + Z_t \indnorm{x}{\invMt}$. Event $E^{\ls}$ concerns the concentration of the MLE and does not depend on the algorithm. By \citet[Theorem~2]{abbasi2011improved}, we have $p_1 \leq 1/T$. By definition of $\ftilde_t$, $\Prob{E^{\anti}_t} = \Prob{Z_t>c_1} \eqqcolon p_3$ and
$\Prob{\overline{E^{\conc}_t}} = \Prob{|Z_t|>c_2}\eqqcolon p_2$. These relations combined with the bound~\cref{genericregretbound} concludes the proof. 
\end{proof}

\vspace{-3ex}
Similar to the MAB case, we choose $U= 3 c_1$ for the default variant of $\randucb$. This ensures  $\Prob{Z > c_1}$ is a positive constant and $\Prob{|Z|> c_2} = 0$, resulting in the promised $\tilde O(d\sqrt{T})$ regret bound. We reiterate that this bound does not have a dependence on $K$, and thus holds in the infinite-arms case. 

\subsubsection{Proof for regret bound for MAB}
\label{sec:linear:mab_proof}
We now present the proof of Theorem~\ref{thm:reg-randucb-mab} by using a reduction from linear bandits to multi-armed bandits. 

Consider a linear bandit (LB) with dimension $d=K$, where arm features correspond to standard basis vectors, i.e., $x_i$ is a one-hot vector with the $i$th component set to 1,

and the true parameter vector is $\thetastar=(\mu(1),\dots,\mu(K))$. Now consider using $\randucb$ with $\lambda=1$ for this problem.

We claim that $\randucb$ for MAB (Eq.\ \eqref{eqn:randucb-mab}) selects the same action on round $t > K$ as $\randucb$ for LB (Eq.~\eqref{eq:randucb-lb}) on round $t-K$. In fact, for any $t>K$, let $s_i(t)$ denote the number of times arm $i$ has been pulled during rounds $K+1,\dots,t$, and so $M_t^{-1}$ is a diagonal matrix with the $i$th diagonal entry $(s_i(t)+1)^{-1}$. $\randucb$ for MAB in round $t$ pulls $i_t = \argmax_i \left\{\muhat_t(i) + Z_t/\sqrt{s_i(t)+1}\right\}$, while $\randucb$ for LB in round $t$ pulls $i_t = \argmax_i \{\dop{x_i}{\thetahat_t}+Z_t\|x_i\|_{M_t^{-1}}\}$. Observe that these expressions are identical, hence $\randucb$ for MAB exactly corresponds to $\randucb$ for LB.

So, Theorem~\ref{thm:linbanditsmainthm} in \cref{sec_reg_linear} applies with $\ftilde_t(x)\coloneqq\dop{x}{\thetahat_t}+Z_t\|x\|_{M_t^{-1}}$.

We next bound the probabilities $p_1$, $p_2$, and $p_3$ define in  Theorem~\ref{thm:linbanditsmainthm}. By Hoeffding's inequality, for each arm $i$ and round $t$:
\begin{align*}
&\Prob{\left|\dop{x_i}{\thetahat_t-\thetastar}\right|>c_1 \|x_i\|_{M_t^{-1}}}
\\&=
\Prob{\left|\muhat_i(t) - \mu_i \right|>\frac{c_1}{{\sqrt{1+s_i(t)}}}}<
1/ KT^2.
\end{align*}
Thus by a union bound over the arms and the rounds, we get $\Prob{\overline{E^{\ls}}} \leq 1/T \eqqcolon p_1$. We can bound $p_2$ and $p_3$  by definition of $\ftilde_t$: we have $\Prob{E^{\anti}_t} = \Prob{Z>c_1}\eqqcolon p_3$ and
$\Prob{\overline{E^{\conc}_t}} = \Prob{|Z|>c_2}\eqqcolon p_2$. Using these bounds together with the bound in~\cref{genericregretbound} completes the proof of Theorem~\ref{thm:reg-randucb-mab}. 
\subsection{Generalized linear bandits}
\label{sec:GLB}
We next consider structured bandits where the feature to reward mapping is a generalized linear model~\citep{mccullagh1984generalized}, meaning that the expected reward in round $t$ satisfies $\E[Y_t | i_t = i] = g(\langle x_i, \sth \rangle)\in[0,1]$, where $g$ is a known, strictly increasing, differentiable function, called the \emph{link} function or the \emph{mean} function. 

If $g(x) = x$, we recover  linear bandits, whereas if $g(x) = 1 / (1 + \exp(-x))$, we get logistic bandits. Assuming arm~$1$ is optimal, the regret is
\(R(T) := \sum_{t = 1}^T  \E\left[ g(x_1, \sth) - g(x_{i_t}, \sth) \right]\) and the {\em effective} gap of arm $i$ is $\Delta_i \coloneqq g(x_1, \sth) - g(x_i, \sth)$. 

\subsubsection{Instantiating \randucb}
\label{sec:GLB-algorithm}
As before, we denote $X_t=x_{i_t}$.
Given previous observations $(X_\ell, Y_\ell)_{\ell = 1}^{t-1}$, the MLE in round $t$ can be computed as~\citep{mccullagh1984generalized}
\(\hth \coloneqq \argmin_{\theta} \sum_{\ell = 1}^{t-1} \left[ Y_{\ell} \langle X_{\ell}, \theta \rangle - b(\langle X_{\ell}, \theta \rangle) \right],
\) where $b$ is a strictly convex function such that its derivative is $g$.

Let $H_t(\theta) \coloneqq \sum_{\ell = 1}^{t-1} g'(\langle X_{\ell}, \theta \rangle) X_{\ell} X_{\ell}\transpose$ denote the Hessian at point $\theta$ on round $t$, and $H_t \coloneqq H_t(\hth)$. We assume that $g$ is $\cL$-Lipschitz, i.e.,
$|g(x)-g(y)| \leq \cL|x-y|$, implying $0<g'(x)\leq \cL$ for all $x$.

Note that in general, matrix $H_t$ is not guaranteed to be positive definite. To guarantee the positive definiteness of $H_t$, we make the following assumptions.\footnote{These assumptions are standard in the analysis of generalized linear bandits~\citep{li2017provable, kveton2019randomized}.} \textbf{(i)} Feature vectors span the $d$-dimensional space. In particular, we assume that there exist basis vectors $\{ v_j \}_{j = 1}^{d} \subseteq \{x_i\}_{i \in \cA}$ with $\sum_{j = 1}^{d} v_j v_j^{T} \succeq \rho  I$ for some $\rho > 0$. This assumption is natural as it would not hold only when the actual dimensionality of the problem is smaller than $d$.
\textbf{(ii)} We assume  
$$\mu \coloneqq  \inf\{ g'(\langle x, \theta \rangle): {\norm{x} \leq 1, \norm{\theta - \sth} \leq 1}\}>0.$$
This assumption holds for all interesting link functions, such as in linear and logistic regression.

$\randucb$ for GLB starts by pulling each of the $v_i$ for 
$O(d \ln(T) / \mu^{2} \rho)$ many times.
We shall show that after this initialization, with probability at least $1-1/T$ we have that
 $\|\thetahat_t-\sth\|\leq1$ and
further $H_t $ is positive-definite for all subsequent rounds.

After this initialization, $\randucb$ follows the same algorithm as for linear bandits (Eq.~\eqref{eq:randucb-lb}),
except that there is no regularization in this case (so, $M_t = \sum_{\ell=1}^{t-1} X_{\ell}X_{\ell}\transpose$).

The corresponding OFU-based algorithm~\citep{li2017provable} has $\beta = \frac{1}{\mu} \sqrt{\frac{d}{2} \, \ln(1 + 2T/d) + \ln(T)}$.
Let $c_1 \coloneqq \sqrt{d\ln(T/d)+2\ln(T)}/2\mu$, and
choose $U= 2 \sqrt{\cL} \, c_1$ for \randucb; the following theorem, proved in \cref{sec_reg_genlinear}, gives the promised $\tilde O(d\sqrt{T})$ regret bound by choosing $c_2 = 3 \sqrt{\cL} \, c_1$.

\begin{theorem}
\label{thm:glb}
Let $c_1 = \sqrt{d\ln(T/d)+2\ln(T)}/2\mu$, $c_3  \coloneqq 2d \ln\left(1 + \frac{T}{d }\right)$. For any $c_2>c_1$, the regret $R(T)$ of $\randucb$ for generalized linear bandits is bounded by
\begin{align*}
&\left(c_1 + \frac{c_2}{\sqrt \mu} \right) \left(1 + \frac{2}{\Prob{Z> c_1\sqrt{\cL}} - \Prob{|Z|> c_2}} \right)  \\ & \times  \cL \sqrt{c_3 T}
+ T \; \Prob{|Z|> c_2} + O(d^2 \ln(T) / \mu^{2} \rho).
\end{align*}
\end{theorem}
\section{Experiments}
\label{sec:experiments}
\begin{figure*}[t]
    \centering
    \subfigure[Various configurations of MAB with large/small gaps (easy/hard) and different reward distributions.]{
    \includegraphics[width = 0.9\textwidth]{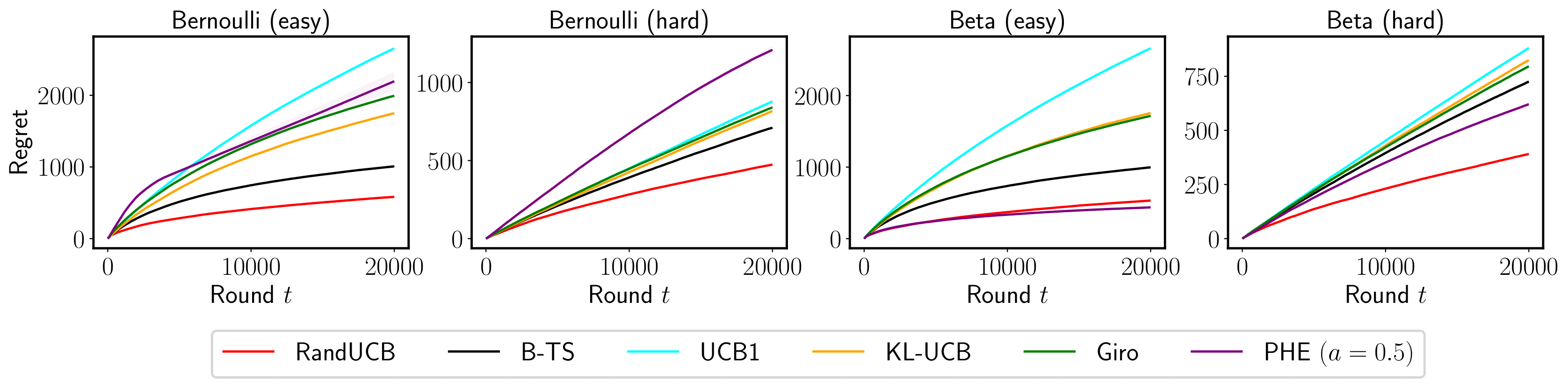}
    \label{fig:results:benchmarking:mab}    
    }
    \subfigure[Linear bandits of different dimensions $d$. RandLinUCB is the instantiation of $\randucb$ for linear bandits.]{
    \includegraphics[width = 0.9\textwidth]{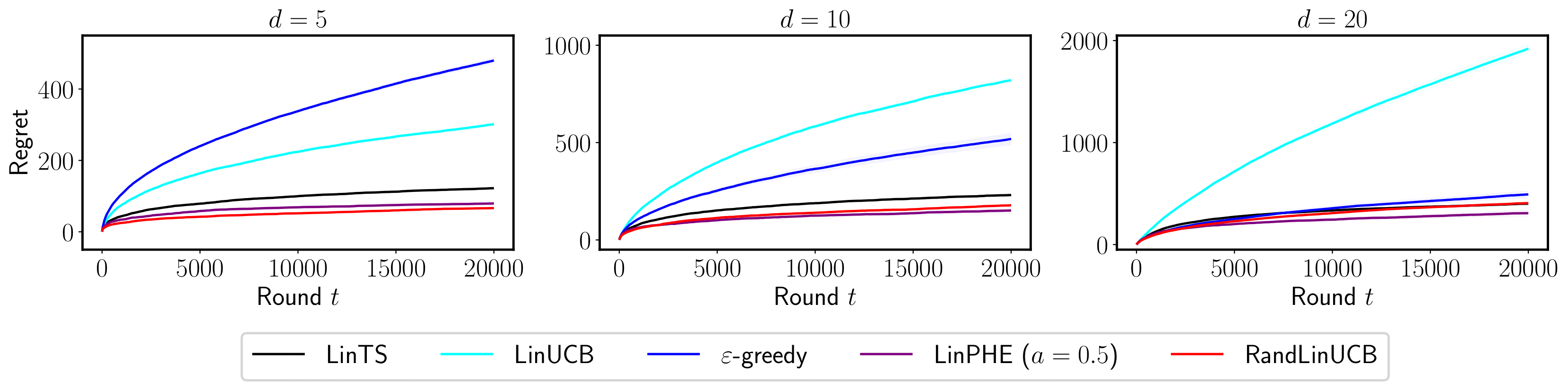}
    \label{fig:results:benchmarking:lin}    
    }
    \subfigure[Logistic bandits of different dimensions $d$. RandUCBLog is the instantiation of $\randucb$ for logistic bandits.]{
    \includegraphics[width = 0.9\textwidth]{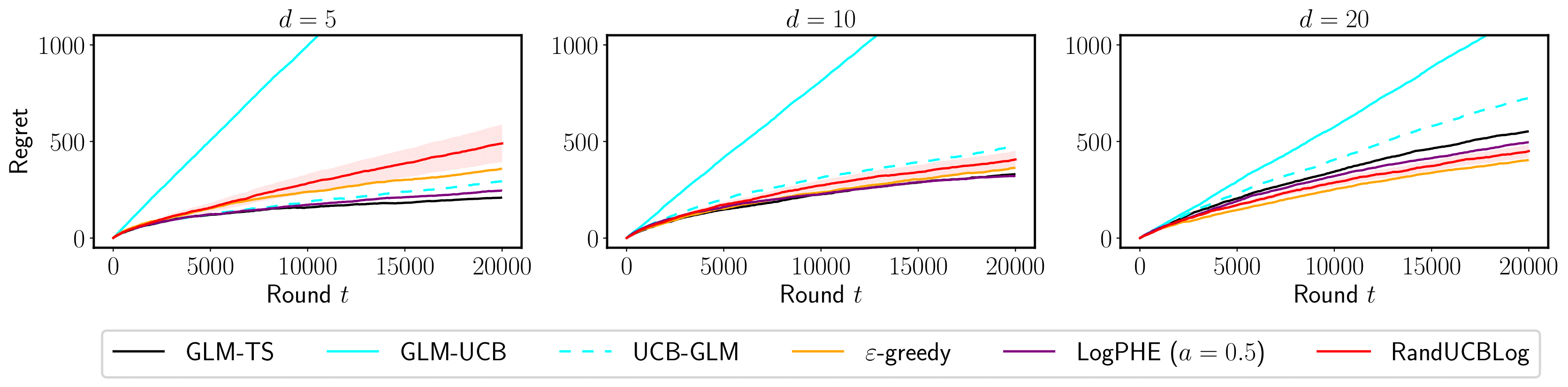}
    \label{fig:results:benchmarking:genlin}
    }
\caption{Cumulative empirical regrets of $\randucb$ versus competitors on various bandit settings.}
\label{fig:results:benchmarking}
\end{figure*}
Finally, we empirically evaluate the performance of $\randucb$ on the bandit settings studied in this paper. We compare various algorithms based on their cumulative empirical regret $\sum_{t=1}^T \left[ Y_t^\star - Y_t \right]$, where  $Y_t^\star$ denotes the reward received by the optimal arm and $Y_t$ is the reward received by the algorithm in round $t$. For all the experiments, we consider $|\cA| = K = 100$ arms and set $T=20,000$ rounds. We average our results over $50$ randomly generated bandit instances.

\paragraph{Multi-armed Bandits:}
\label{sec:MAB-exps}
We first consider the MAB setting and investigate the impact of the gap sizes and the reward distribution. We consider an easy class and a hard class of problem instances:
in the former, arm means are sampled uniformly in $[0.25,0.75]$, while in the latter, they are sampled in $[0.45,0.55]$. We consider both discrete, binary rewards sampled from Bernoulli distributions, as well as continuous rewards sampled from beta distributions. We present results for a Gaussian reward distribution in~\cref{app:expes:gaussian}. In \cref{app:expes:ablation}, we investigate the impact of the design choices and parameters of $\randucb$ in the MAB setting. Recall that $\randucb$ is characterized by the choice of sampling distribution (\cref{sec:randucb:sampling_dist}). We compare the performance of the uniform and Gaussian distributions (with different standard deviations $\sigma$), and observe in Figure~\ref{fig:results:ablation:sampling_distribution} that lower values of $\sigma$  result in better  performance in all our experiments. We also observe in Figure~\ref{fig:results:ablation:nbins} that $\randucb$ is robust to the value of $M$,  the extent of discretization. Note that previous work has also observed that the empirical performance of UCB1 can be improved by using smaller confidence intervals than suggested by theory~\citep{hsu2019empirical, li2012unbiased}, e.g., by tuning $\beta$.
In contrast to our work, these heuristics do not have theoretical guarantees. 

We then estimate the impact of optimism and coupling of the arms on the empirical performance of \randucb. In Figure~\ref{fig:results:ablation:optimism_coupling}, we observe that coupling the arms is more determinant in improving the performance of $\randucb$ compared with optimism, which has only a minor effect. We notice that this phenomenon is also observed for TS: the optimistic variant of TS (OTS) has similar performance to TS (Figure~\ref{fig:results:optimistic_ts} in \cref{app:expes:ots}). 

Following the above ablation study, in the following experiments we initiate $\randucb$ with a (discretized, optimistic) Gaussian sampling distribution and coupled arms with parameters $\epsilon=10^{-7}$, $\sigma=1/8$, $L=0$, $U=2\sqrt{\ln(T)}$, and $M=20$. Figure~\ref{fig:results:benchmarking:mab} compares $\randucb$ against classical and state-of-the-art baselines. 
In particular, we compare against TS with Bernoulli-Beta conjugate priors (B-TS)~\citep{agrawal2013further} and UCB1~\citep{auer2002finite}. We also consider the much tighter KL-UCB version~\citep{garivier2011kl}, in addition to the recent GiRo~\citep{kveton2019garbage} and PHE~\citep{kveton2019perturbedmab} algorithms 
and observe that $\randucb$ performs consistently well, clearly outperforming all baselines in three settings, while matching the performance of PHE in the remaining setting. Most importantly, it outperforms TS in all settings.

\paragraph{Structured Bandits:}
\label{sec:structured-exps}
For structured bandits, we use the same setting of $\randucb$ described above but with the confidence intervals given by the specific bandit problem. We consider linear bandits as well as logistic regression for the generalized linear case. For each of these problems, we vary the dimension $d\in\{5, 10, 20\}$. Each problem is characterized by an (unknown) parameter  $\theta^\star$ and $K$ arms. We consider Bernoulli $\{0,1\}$ rewards.\footnote{To make sure the expected rewards lie in $[0,1]$, we choose each of $\sth$ and the feature vectors by
sampling a uniformly random $(d-1)$-dimensional vector of norm $1/\sqrt2$ and concatenate it with a $1/\sqrt2$ component.}

For $\randucb$ in the linear bandit setting, we use the same hyper-parameters as before, but set $U = \beta = \sqrt{\lambda}+\frac12\sqrt{\ln (T^2\lambda^{-d} \det(M_t) )}$, which is the value
from the corresponding OFU-based algorithm~\citep[Theorem~2]{abbasi2011improved}, and $\lambda = 10^{-4}$. For comparison, we consider two variants of LinTS~\citep{abeille2017linear, agrawal2013thompson}: a theoretically optimal variant with the covariance matrix ``inflated'' by a dimension-dependent quantity and the more commonly used variant without this additional inflation~\citep{chapelle2011empirical}. We also consider LinUCB~\citep{abbasi2011improved},  $\epsilon$-greedy~\citep{langford2008epoch}, and the best performing variant of the randomized strategy LinPHE~\citep{kveton2019perturbedlinear}. For $\epsilon$-greedy, we chose the best performing value of $\epsilon=0.05$ and anneal it as $\epsilon_t = \frac{\epsilon \sqrt{T}}{2 \sqrt{t}}$. 

For $\randucb$ in the GLB setting, we use the same hyper-parameters as before, but now set $U = \beta= \frac{1}{\mu} \sqrt{\frac{d}{2} \, \ln(1 + 2T/d) + \ln(T)}$, which is the constant
from the corresponding OFU-based algorithm~\citep{li2017provable}. We compare against GLM-TS~\citep{abeille2017linear, kveton2019randomized}, which samples from a Laplace approximation of the posterior distribution. We consider two OFU-based algorithms: GLM-UCB~\citep{filippi2010parametric} and UCB-GLM~\citep{li2017provable}. For \randucb, we chose to randomize the tighter confidence intervals in UCB-GLM by the same scheme in Eq.~\eqref{eq:randucb-lb}. We further compare against $\epsilon$-greedy~\citep{langford2008epoch} and the best performing variant of LogPHE~\citep{kveton2019randomized}.

Figure~\ref{fig:results:benchmarking:lin} shows that $\randucb$ matches the performance of the best strategies in linear bandits. Figure~\ref{fig:results:benchmarking:genlin} shows that $\randucb$ is competitive against other state-of-the-art strategies in logastic bandits. These results confirm that $\randucb$ is robust to the problem configuration and is an effective randomized alternative to  complicated strategies.
\section{Conclusion}
\label{sec:conclusion}
We introduced the $\randucb$ meta-algorithm as a generic strategy for randomizing OFU-based algorithms. Our results across bandit settings illustrate that $\randucb$ matches the empirical performance of TS (and often outperforms it) and yet attains the theoretically optimal regret bounds of  OFU-based algorithms, thus achieving the best of both worlds. An additional advantage of $\randucb$ is its broad applicability: the same mechanism of randomizing upper confidence bounds can be potentially used to improve the performance of other OFU-based algorithms. This could be useful in domains such as Monte-Carlo tree search~\citep{kocsis2006bandit} and risk-aware bandits~\citep{galichet2013exploration}, where designing randomized exploration strategies is not straightforward, as well as for practical scenarios such as  delayed rewards~\citep{chapelle2011empirical}, where randomization is crucial for robustness. 

\clearpage
\section{Acknowledgements}
We would like to thank Aditya Ramdas for pointing out a mistake in an earlier version of the paper. This research was partially supported by an IVADO postdoctoral scholarship. 
Abbas Mehrabian was supported by an IVADO-Apog\'ee-CFREF postdoctoral fellowship.
\bibliographystyle{plainnat}
\bibliography{ref}

\clearpage
\onecolumn
\appendix
\section{Regret Bound for Multi-Armed Bandits with Uncoupled Arms} 

\begin{proposition}[Hoeffding's inequality
\protect{\citep[Theorem~2]{hoeffding}}]
Let $X_1,\dots,X_n$ be independent random variables taking values in $[0,1]$. Then, for any $t\geq0$,
\[
\mathbb{P} \left[
\sum_{i=1}^{n} (X_i - \mathbb{E}X_i) > tn\right] <  \exp (-2nt^2).
\]
\end{proposition}

Recall that $\randucb$ first pulls each arm once and then in round $t>K$, chooses an arm $i$ maximizing
$\muhat_i(t) 
+ \frac{Z_{i,t}}{\sqrt{s_i(t)}}$, where the $Z_{i,t}$ are i.i.d.\ and distributed like a given random variable $Z$. We may assume, without loss of generality, that arm 1 is the unique optimal arm.
For a random variable $X$, the notation $\P_{X}$ means taking the probability with respect to the randomness in $X$.
We will also use the shorthand $\muhat_{i,s}\coloneqq \muhat_i(s)$.

We will use the following result, which follows from Theorem~1 from \citet{kveton2019garbage}.
\begin{theorem}\label{girotheorem}
Let
$\tau_2,\dots,\tau_K$ be arbitrary but deterministic. The regret of $\randucb$ after $T$ rounds can be upper bounded by $K+\sum_{i=2}^{K} \Delta_i (a_i+b_i)$,
where
\begin{align*}
a_i & \coloneqq \sum_{s=1}^{T-1}
\E_{\muhat_{1,s}}
\left[
\min\left\{\frac{1}{\P_{Z_{1,s}}\left(\muhat_1(s) 
+ \frac{Z_{1,s}}{\sqrt{s}}\geq\tau_i\right)}-1,T\right\}
\right],\\
b_i & \coloneqq
1 + \sum_{s=1}^{T-1}
\P_{\muhat_{i,s}} \left\{
\P_{Z_{i,s}}\left\{
\muhat_i(s) 
+ \frac{Z_{i,s}}{\sqrt{s     }}\geq \tau_i
\right\} > 1/T
\right\}.
\end{align*}
\end{theorem}

In the rest of this section, we explain how the following 
result follows from the above theorem.
Assume that $Z$ has a discrete distribution and takes value $\alpha_i$ with probability $p_i$.

\label{sec:instancedependentupperbound}
\begin{theorem}\label{thm:rucbmainregbound}
Assume that
$0\leq\alpha_1 \leq \alpha_2\leq \cdots \leq \alpha_M$,
 and
$p_i\geq0$,
and  $\sum p_i=1$, and suppose $p_M>1/T$.
Then, the regret of $\randucb$ after $T$ rounds is bounded by
\begin{align*}
K+\left(\sum_{n=1}^{M-1}
\left(\frac{p_1+\dots+p_{n}}
{p_{n+1} + \dots + p_M}\right)
e^{-2\alpha_n^2}
+T e^{-2\alpha_M^2}
+4+
3 \alpha_M^2 
\right)\cdot\left(
\sum_{i=2}^{K}
\left(\frac{6}{\Delta_i}\right)\right).
\end{align*}
\end{theorem}
\Cref{thm:instancedependent} follows from Theorem~\ref{thm:rucbmainregbound}
by crudely bounding $\frac{p_1+\dots+p_{n}}{p_{n+1} + \dots + p_M}
\leq 1/p_M$ and $e^{-2\alpha_n^2}\leq1$.

Theorem~\ref{thm:rucbmainregbound} follows from Theorem~\ref{girotheorem}
by setting $\tau_i = \mu_i + \Delta_i/2 = \mu_1 - \Delta_i/2$.
We bound the $a_i$ and the $b_i$ in the following two sections.

\subsection{Bounding the $a_i$}
Let us define
\[
a_{i,s}
\coloneqq 
\E_{\muhat_{1,s}}
\left[
\min\left\{\frac{1}{\P_{Z_{1,s}}(\muhat_1(s) 
+ \frac{Z_{i,s}}{\sqrt{s     }}\geq \tau_i)}-1,T\right\}
\right].
\]
Fix $s$ and for each $1\leq j\leq M$, define the event $E_j$ as 
\[
E_j \coloneqq \left \{ \muhat_{1,s} + \alpha_j/\sqrt{s} \geq \tau_i \right\},
\]
which is deterministic given the history.
Note that $E_M \supseteq E_{M-1} \supseteq \cdots \supseteq E_1$.
Then, define
\[
N \coloneqq \min \{j: E_j \textnormal{ holds}\},
\]
and $N=M+1$ if none of the $E_j$ hold.
Since the events
$N=1, N=2, \dots, N=M, N=M+1$ partition the space, we can bound $a_{i,s}$ as
\begin{align*}
a_{i,s}
& =
\sum_{n=1}^{M+1}
\P_{\muhat_{1,s}}\{N=n\}
\E_{\muhat_{1,s}}
\left[
\min\left\{\frac{1}{\P_{Z_{1,s}}(\muhat_1(s) 
+ \frac{Z_{1,s}}{\sqrt{s     }}\geq \tau_i)}-1,T\right\}
\bigg | N=n
\right]\\
& \leq
\sum_{n=1}^{M}
\P_{\muhat_{1,s}}\{N=n\}
\E_{\muhat_{1,s}}
\left[
\frac{1}{\P_{Z_{1,s}}(\muhat_1(s) 
+ \frac{Z_{1,s}}{\sqrt{s     }}\geq \tau_i)}-1
\bigg | N=n
\right]
+
T \P_{\muhat_{1,s}}\{N=M+1\}.
\end{align*}
Next, observe that by definition of $N$, under the event $N=n$ with $2\leq n \leq M$,
we have
\[
\frac{1}{\P_{Z_{1,s}}(\muhat_1(s) 
+ \frac{Z_{1,s}}{\sqrt{s     }}\geq \tau_i)}-1
=
\frac{1}{p_n+p_{n+1}+\dots+p_M}-1
=
\frac{p_1+\dots+p_{n-1}}
{p_n + \dots + p_M}.
\]
Moreover, if $N=1$ then
$\P_{Z_{1,s}}(\muhat_1(s) 
+ \frac{Z_{1,s}}{\sqrt{s     }}\geq \tau_i)=1$
and thus
$\frac{1}{
\P_{Z_{1,s}}(\muhat_1(s) 
+ \frac{Z_{1,s}}{\sqrt{s     }}\geq \tau_i)}-1
=0$.
Therefore, we have
\[
a_{i,s} \leq \sum_{n=2}^{M}
\P_{\muhat_{1,s}}\{N=n\}
\cdot \left(\frac{p_1+\dots+p_{n-1}}
{p_n + \dots + p_M}\right)
+
T \P_{\muhat_{1,s}}\{N=M+1\}.
\]
We next bound the probability
$\P_{\muhat_{1,s}}\{N=n\}$, for any $2\leq n\leq M+1$.
Note that $N=n$ implies $E_{n-1}$ did not happen. That is,
if $N=n$, then
$$\muhat_{1,s}+\frac{\alpha_{n-1}}{\sqrt s} < \tau_1 = \mu_1 - \Delta_i/2,$$
which is equivalent to
\[
\mu_1 - \muhat_{1,s} > \frac{\alpha_{n-1}}{\sqrt s} + \Delta_i/2.
\]
\notetoself{handle the plus 1}
{Since $\alpha_{n-1}\geq0$, }
we apply Hoeffding's inequality to conclude
\[
\P_{\muhat_{1,s}} \{ N= n\}
\leq
\P_{\muhat_{1,s}} \{ \mu_1 - \muhat_{1,s} > \frac{\alpha_{n-1}}{\sqrt s} + \frac{\Delta_i}{2} \}
\leq
\exp \left( 
-2s (\frac{\alpha_{n-1}}{\sqrt s}+\frac{\Delta_i}{2})^2
\right)
\leq
\exp \left( 
-2 \alpha_{n-1}^2
- s {\Delta_i}^2/2
\right).
\]
Therefore, we have
\[
a_{i,s} \leq \sum_{n=2}^{M}
\exp(-2\alpha_{n-1}^2-s\Delta_i^2/2)
\cdot \left(\frac{p_1+\dots+p_{n-1}}
{p_n + \dots + p_M}\right)
+
T 
\exp(-2\alpha_M^2-s\Delta_i^2/2),
\]
and so
\begin{align*}
a_i
& \leq \sum_{s=0}^{T-1}
\left\{
\sum_{n=1}^{M-1}
\exp(-2\alpha_{n}^2-s\Delta_i^2/2)
\cdot \left(\frac{p_1+\dots+p_{n}}
{p_{n+1} + \dots + p_M}\right)
+
T 
\exp(-2\alpha_M^2-s\Delta_i^2/2)
\right\}\\
&\leq
\sum_{n=1}^{M-1}
\left\{
\left(\frac{p_1+\dots+p_{n}}
{p_{n+1} + \dots + p_M}\right)
e^{-2\alpha_n^2}
\sum_{s=0}^{\infty}
e^{-s\Delta_i^2/2}
\right\}
+
T e^{-2\alpha_M^2}
\sum_{s=0}^{\infty}
e^{-s\Delta_i^2/2}.
\end{align*}
We now bound
\begin{equation}
\sum_{s=0}^{\infty} e^{-s\Delta_i^2/2}
= \frac{1}{1-e^{-\Delta_i^2/2}}
\leq
\frac{1}{\Delta_i^2/6},
\label{exponentialsum}
\end{equation}
which gives
\[
a_i \leq
\left(\frac{6}{\Delta_i^2}\right) \cdot
\left(\sum_{n=1}^{M-1}
\left(\frac{p_1+\dots+p_{n}}
{p_{n+1} + \dots + p_M}\right)
e^{-2\alpha_n^2}
+
T e^{-2\alpha_M^2}\right).
\] 

\subsection{Bounding the $b_i$}
Recall that
\[b_i -1 =
\sum_{s=0}^{T-1}
\P_{\muhat_{i,s}} \left\{
\P_{Z_{i,s}}\left\{
\muhat_{i,s}+\frac{Z_{i,s}}{\sqrt{s}}\geq \tau_i
\right\} > 1/T
\right\}.\]
Let 
$$b_{i,s}\coloneqq
\P_{\muhat_{i,s}} \left\{
\P_{Z_{i,s}}\left\{
\muhat_{i,s}+\frac{Z_{i,s}}{\sqrt{s}}\geq \tau_i
\right\} > 1/T
\right\}.
$$
By the definition of $Z_{i,s}$, we have
\[
b_{i,s}=
\P_{\muhat_{i,s}} \left\{
\sum_{n=1}^{M}
p_n \ind {\muhat_{i,s}+\frac{\alpha_n}{\sqrt s} \geq \tau_i} > 1/T
\right\}.
\]
Since $p_M>1/T$, we have
\[
b_{i,s}=
\P_{\muhat_{i,s}} \left\{
\muhat_{i,s}+\frac{\alpha_M}{\sqrt s} \geq \tau_i
\right\}=
\P_{\muhat_{i,s}} \left\{
\muhat_{i,s}+\frac{\alpha_M}{\sqrt s} \geq \mu_i+\Delta_i/2
\right\}.
\]
For
$s\leq \lfloor 16 \alpha_M^2/\Delta_i^2 \rfloor $,
we simply upper bound $b_{i,s}\leq1$, while
for
$s\geq \lceil 16 \alpha_M^2/\Delta_i^2 \rceil$,
we have 
$\Delta_i/2-\alpha_M/\sqrt{s}
\geq\Delta_i/4>0$, 
so
by Hoeffding's inequality,
\[
b_{i,s}
\leq \P_{\muhat_{i,s}}
\left\{
\muhat_{i,s}-\mu_i \geq \Delta_i/4
\right\}
\leq
\exp(-s \Delta_i^2/8).
\]
Therefore,
\begin{align*}
b_{i}-1 
=
\sum_{s=0}^{T-1} b_{i,s}
\leq
\lfloor 16 \alpha_M^2/\Delta_i^2 \rfloor
+
\sum_{s=\lceil 16 \alpha_M^2/\Delta_i^2 \rceil}^{T-1} b_{i,s}
\leq 16 \alpha_M^2/\Delta_i^2
+ \sum_{s=0}^{\infty}
e^{-s\Delta_i^2/8}
\leq
\frac{16 \alpha_M^2 + 24}{\Delta_i^2},
\end{align*} 
where we have used
\(
\sum_{s=0}^{\infty} e^{-s\Delta_i^2/8}
\leq
\frac{24}{\Delta_i^2},
\)
proved as in~\eqref{exponentialsum}.
\section{A regret bound for a general index-based algorithm for linear bandits}
\label{sec_reg_linear}
In this appendix, we obtain an analogous result as Theorem~1 from~\citet{kveton2019perturbedlinear} for any index-based algorithm for linear bandits which
in each round $t$ selects the arm $i_t = \argmax_{i} \tilde{f}_{t}(x_i)$, where $\tilde{f}_{t}$ is a stochastic function depending on the history up to round $t$, and possibly additional independent randomness. 
In $\randucb$, we will set $\tilde{f}_{t}(x) \coloneqq \langle \hth, x \rangle + Z_t \indnorm{x}{\invMt}$, where $Z_1,\dots,Z_T$ are i.i.d.\ random variables. 
The proof is quite similar to that of Theorem~1 from \citet{kveton2019perturbedlinear}, where this result was proved for the special case when $\ftilde_t$ is linear.
To extend the proof to a general $\ftilde_t$, the first step in the proof of Lemma~\ref{lem:inst_regret} has been changed. We include the complete proof for completeness.

\begin{theorem}
[Generic regret bound for index-based algorithms for linear bandits]
\label{thm:linbanditsmainthm}
Let 
\(c_3  \coloneqq 2d \ln\left(1 + \frac{T}{d \lambda}\right)
\).
Suppose $c_1<c_2$ are real numbers satisfying $c_1+c_2\geq1$ and define the events
\begin{align*}
E^{\ls} & \coloneqq \left\{ \forall i \in \cA,\forall
d+1\leq t \leq T
\text{;} \quad \vert \langle x_i, \hth - \sth \rangle 
\vert \leq c_1 \indnorm{x_i}{\invMt} \right\},\\
E^{\conc}_{t} & \coloneqq \left\{ \forall i \in \cA \text{;} \quad \vert \tilde{f}_{t}(x_i) - \langle x_i, \hth  \rangle \vert \leq c_2 \indnorm{x_i}{\invMt} \right\}, \text{and}\\
E^{\anti}_{t} &\coloneqq \left\{ \tilde{f}_{t}(x_1) - \langle x_1, \hth  \rangle > c_1 \indnorm{x_1}{\invMt} \right\}.
\end{align*} 
Suppose $p_1,p_2,p_3\in[0,1]$ are such that
$\Prob{E^{\ls}} \geq 1-p_1$
and that for each given $ t \leq T$ and \emph{for any possible history $\cH_{t-1}$ until the end of round $t-1$,}
we have
\[
\Pr(E^{\anti}_t|\cH_{t-1}) \geq p_3,
\textnormal{ and }
\Pr(E^{\conc}_t|\cH_{t-1}) \geq 1-p_2.
\]
Then, the regret after $T$ rounds is bounded by
\(
 (c_1 + c_2) \left(1 + \frac{2}{p_3 - p_2} \right) \sqrt{c_3 T} + T \; (p_1 + p_2).  
\)
\end{theorem}

In the rest of this section, we prove Theorem~\ref{thm:linbanditsmainthm}. 
Let use denote by $\E_t,\Pr_t$ the randomness injected by the algorithm in round $t$ (i.e., the randomness of $Z_t$),
and denote by $\E_{\cH_{t-1}},\Pr_{\cH_{t-1}}$ the randomness in the history.
The following lemma is an analogue of Lemma 2 from \citet{kveton2019perturbedlinear}.
Let  $\Delta_i \coloneqq \dop{x_1-x_i}{\thetastar}$ denote the gap of arm $i$.

\begin{lemma}
\label{lem:inst_regret}
For any round $t$ and any history $\cH_{t-1}$, we have
\[
\E_t[\Delta_{i_t} \ind{E^{\ls}} | \cH_{t-1}]\leq p_2+(c_1+c_2) \left( 1+\frac{2}{p_3-p_2}\right)
\E_t[\min\{1,\|X_t\|_{M_t^{-1}}\} | \cH_{t-1}].
\]
\end{lemma}
Before proving this lemma we show how it implies Theorem~\ref{thm:linbanditsmainthm}.
We will be using Lemma~11 from \citet{abbasi2011improved}, which states that, deterministically,
\begin{equation}
\sum_{t=1}^T \min\{1,\|X_t\|_{M_t^{-1}}\}^2 \leq c_3.
\label{19lemma3}
\end{equation}

\begin{proof}[Proof of Theorem~\ref{thm:linbanditsmainthm}]
Recall that
$X_t= x_{i_t}$ and $Y_t$ denotes the reward received in round $t$.
By Eq.~\eqref{19lemma3} 
and the Cauchy-Schwarz inequality, we have, deterministically,
\[
\sum_{t=1}^T \min\{1,\|X_t\|_{M_t^{-1}}\} \leq \sqrt{c_3 T}.
\]
Thus, since $\Delta_i\leq 1$ for all $i$, and using Lemma~\ref{lem:inst_regret} we have \todob{Is there any standard notation for random variables? I suggest that we use capital letters. So $i_t$ would be $I_t$.}
\abas{we don't have a consistent notation now, but let's do this later because it's  easy  to change}
\begin{align*}
\textnormal{Regret} & 
=        \sum_{t=  1}^T \E \Delta_{i_t}\\
&\leq     T \P (\overline{E^{\ls}})+ \sum_{t=  1}^T \E [\Delta_{i_t}\ind{E^{\ls}}]\\
&\leq     T p_1 + \sum_{t=1}^{T}
\E_{\cH_{t-1}} [ \E_t [\Delta_{i_t}\ind{E^{\ls}} | \cH_{t-1}] ]\\
&\leq     T p_1 + \sum_{t=  1}^{T}
\E_{\cH_{t-1}} \left[ p_2+(c_1+c_2) \left( 1+\frac{2}{p_3-p_2}\right)
\E_t[\min\{1,\|X_t\|_{M_t^{-1}}\} | \cH_{t-1}] \right]\\
&=     T p_1 + \sum_{t=  1}^{T}
\E \left[ p_2+(c_1+c_2) \left( 1+\frac{2}{p_3-p_2}\right)
\min\{1,\|X_t\|_{M_t^{-1}}\} \right]\\
&=     T p_1 + Tp_2+(c_1+c_2) \left( 1+\frac{2}{p_3-p_2}\right)
\E \sum_{t=  1}^{T} \min\{1,\|X_t\|_{M_t^{-1}}\}  \\
&\leq     T (p_1 + p_2)+(c_1+c_2) \left( 1+\frac{2}{p_3-p_2}\right)
\sqrt{c_3 T},
\end{align*}
completing the proof of the theorem.
\end{proof}

We next prove Lemma~\ref{lem:inst_regret}.

\begin{proof}[Proof of Lemma~\ref{lem:inst_regret}]
We fix a round $t$ and an arbitrary history $\cH_{t-1}$ satisfying $E^{\ls}$, and omit the conditioning on $\cH_{t-1}$ henceforth.
Let us denote $N\coloneqq M_t^{-1}$.
Let $c\coloneqq c_1+c_2$ and define
\[
S_t \coloneqq \{i\in A : c \|x_i\|_{N} < \Delta_i\}
\textnormal{ and }
\overline{S_t} = A \setminus S_t.
\]
Observe that $S_t$ is deterministic (since we have fixed the history) and that $1 \notin S_t$.
The arms in $S_t$ are sufficiently sampled, and the rest of the arms are undersampled.
Also, let 
\[
j_t \coloneqq \argmin_{i\notin S_t} \|x_i\|_{N}
\]
be the least uncertain undersampled arm, which is deterministic given the history. We may write
\begin{align}
\E [\Delta_{i_t}] &=
\E [\Delta_{i_t} \ind{E^{\conc}_t}]
+ \E [\Delta_{i_t} \ind{\overline{E^{\conc}_t}}] \notag\\
&\leq
\E [\Delta_{i_t} \ind{E^{\conc}_t}]
+ \Prob{\overline{E^{\conc}_t}} \notag\\
&\leq
\E [\Delta_{i_t} \ind{E^{\conc}_t}]
+ p_2\label{deltapartition}.
\end{align}
Recall that $E^{\conc}_t$ is the event that for all arms $i$,
$|\ftilde_t(x_i) - \dop{x_i}{\thetahat_t}|\leq c_2 \|x_i\|_N$,
and $E^{\ls}$ is the event that for all arms $i$,
$| \dop{x_i}{\thetahat_t-\thetastar}|\leq c_1 \|x_i\|_N$.
Hence, on event $E^{\conc}_t\cap E^{\ls}$ we have
\[
\ftilde_t(x_{i_t})
\leq c_2 \|x_{i_t}\|_N + 
\dop{x_{i_t}}{\thetahat_t}
\leq c_2 \|x_{i_t}\|_N + 
c_1 \|x_{i_t}\|_N +
\dop{x_{i_t}}{\theta^*},
\]
and similarly,
\[
\ftilde_t(x_{j_t}) \geq 
\dop{x_{j_t}}{\theta^*} - c \|x_{j_t}\|_N,
\]
which, since $\ftilde_t(x_{i_t})\geq
\ftilde_t(x_{j_t})$, gives
\[\Delta_{i_t}=\Delta_{j_t}+
\dop{x_{j_t}-x_{i_t}}{\thetastar}
 \leq c \|x_{j_t}\|_N + (c \|x_{i_t}\|_N+c \|x_{j_t}\|_N)
 = c( \|x_{i_t}\|_N+ 2\|x_{j_t}\|_N)
\]
deterministically.
On the other hand, since $\Delta_{i_t}\leq1\leq c$, we also have
\[
\Delta_{i_t}\leq c( (1\wedge\|x_{i_t}\|_N)+ 2(1\wedge\|x_{j_t}\|_N)),
\]
where $a \wedge b \coloneqq \min\{a,b\}$.
Plugging into Eq.~\eqref{deltapartition}, we obtain
\begin{equation}\label{expecteddelta}
\E[\Delta_{i_t}] \leq 
c( \E(1\wedge\|x_{i_t}\|_N)+ 2(1\wedge\|x_{j_t}\|_N))+p_2.
\end{equation}
The next step is to bound $1\wedge\|x_{j_t}\|_N$ from above.
Observe that,
\[
\E(1\wedge\|x_{i_t}\|_N)
\geq
\Ex{(1\wedge\|x_{i_t}\|_N) | i_t \in \overline{S_t}}\Prob{i_t \in \overline{S_t}}
\geq
(1\wedge\|x_{j_t}\|_N) \Prob{i_t \in \overline{S_t}},
\]
where the last inequality is by the definition of $j_t$.
Rearranging gives
\[
1\wedge\|x_{j_t}\|_N
\leq
\E(1\wedge\|x_{i_t}\|_N)
\big/\Prob{i_t \in \overline{S_t}}.
\]
Next, we bound $\Prob{i_t \in \overline{S_t}}$ from below.
By definition of $i_t$ and since $1\in\overline{S_t}$, we have
\[
\Prob{i_t \in \overline{S_t}}
\geq
\Prob{\ftilde_t(x_1) > \max_{j\in S_t}\ftilde_t(x_j)}
\geq
\Prob{\ftilde_t(x_1) > \max_{j\in S_t}\ftilde_t(x_j) \textnormal{ and } E^{\conc} \textnormal{ holds}}.
\]
If $E^{\conc}$ holds, then for any $j\in S_t$ we have
\[
\ftilde_t(x_j) \leq \dop{x_j}{\thetastar} + c \|x_j\|_N
<
\dop{x_j}{\thetastar} + \Delta_j
= \dop{x_1}{\thetastar},
\]
whence,
\begin{align*}
\Prob{\ftilde_t(x_1) > \max_{j\in S_t}\ftilde_t(x_j) \textnormal{ and } E^{\conc} \textnormal{ holds}}
& \geq
\Prob{\ftilde_t(x_1) >  \dop{x_1}{\thetastar} \textnormal{ and } E^{\conc} \textnormal{ holds}}\\
&\geq
\Prob{\ftilde_t(x_1) >  \dop{x_1}{\thetastar}}
- \Prob{\overline{E^{\conc}}}.
\end{align*}
Finally, note that if $E^{\anti}\cap E^{\ls}$ holds, then
\[
\ftilde_t(x_1)> \dop{x_1}{\thetahat_t}+c_1\|x_1\|_N
\geq \dop{x_1}{\thetastar},
\]
and thus
\[
\Prob{\ftilde_t(x_1) >  \dop{x_1}{\thetastar}}
- \Prob{\overline{E^{\conc}}}
\geq p_3-p_2.
\]
Hence, we find
$\Prob{i_t \in \overline{S_t}}\geq p_3-p_2$, and plugging this back into Eq.~\eqref{expecteddelta} gives
\[
\E [\Delta_{i_t}] \leq 
c (\E (1\wedge\|x_{i_t}\|_N) + 2 (1\wedge\|x_{j_t}\|_N))+p_2
\leq
c (\E (1\wedge\|x_{i_t}\|_N) + \frac{2 \E(1\wedge\|x_{i_t}\|_N)}{p_3-p_2})+p_2,
\]
completing the proof of the lemma.
\end{proof}
\section{Regret bounds for general index-based algorithms for generalized linear bandits}
\label{sec_reg_genlinear}

In this section, we prove Theorem~\ref{thm:glb}.
Recall that we denote by $i_t$ the arm pulled in round $t$. Let $X_t\coloneqq x_{i_t}$ and let $Y_t$ denote the reward received in round $t$. We define $\hth$ as the MLE for the generalized linear model (GLM) at round $t$. We denote the Hessian matrix by
$H_t \coloneqq \sum_{\ell = 1}^{t-1} g'(\langle X_{\ell}, \thetahat_t \rangle) X_{\ell} X_{\ell}\transpose$. We also define the Gram matrix $M_t \coloneqq \sum_{\ell=1}^{t-1} X_{\ell}X_{\ell}\transpose$. 
For matrices $A$ and $B$, by $A\succeq B$ we mean that $A-B$ is positive semidefinite,
and by $A\succ B$ we mean that $A-B$ is positive definite.
Recall that
$\{v_j\}_{j = 1}^{d} \subseteq \{x_i\}_{i \in\cA}$ forms a basis such that $\sum_{j = 1}^{d}  v_j v_j\transpose \succeq \rho I$.

Similar  to the linear bandit case, we first obtain an analogous result as Theorem~1 from \citet{kveton2019randomized} for any index-based algorithm that does some initialization, then in each subsequent round $t$, selects the arm $i_t = \argmax_{i} \widetilde{f}_{t}(x_i)$, where $\widetilde{f}_{t}$ is a stochastic function depending on the history up to round $t$, and possibly additional independent randomness. For \randucb, the initialization is pulling the $v_j$ vectors sufficiently many times and
we will set $\widetilde{f}_{t}(x) \coloneqq \langle \hth, x \rangle + Z_t \indnorm{x}{\invMt}$, where $Z_1,\dots,Z_T$ are i.i.d.\ random variables.
Recall that $\cL$ is the Lipschitz constant of $g$, and that
$$0<\mu =  \inf\{ g'(\langle x, \theta \rangle): {\norm{x} \leq 1, \norm{\theta - \sth} \leq 1}\}.$$

\begin{theorem}
[Generic regret bound for index-based algorithms for generalized linear bandits]
\label{thm:genlinbanditsmainthm}
Let 
\(c_3  \coloneqq 2d \ln\left(1 + \frac{T}{d}\right).
\)
Suppose $c_1<c_2$ and $\tau$ are real numbers and define the events
\begin{align*}
E^{\mle} & \coloneqq \left\{ \forall i \in \cA,\forall
\tau  < t
\text{:} \quad \vert \langle x_i, \hth - \sth \rangle 
\vert \leq c_1 \indnorm{x_i}{\invMt} \right\}, \\
E^{\bound} &\coloneqq \left\{ \forall \tau<t : \norm{\hth - \sth} \leq 1
\textnormal{ and }
M_t \succeq I
\textnormal{ and }
H_t \succ 0
\right\}, \\
E^{\conc}_{t} &\coloneqq \left\{ \forall i \in \cA \text{:} \quad \vert \widetilde{f}_{t}(x_i) - \langle x_i, \hth  \rangle \vert \leq c_2 \indnorm{x_i}{\invHt} \right\}, \\
E^{\anti}_{t} &\coloneqq \left\{ \widetilde{f}_{t}(x_1) - \langle x_1, \hth  \rangle > \sqrt{\cL} \; c_1 \indnorm{x_1}{\invHt} \right\}.
\end{align*}
Suppose $p_1,p_2,p_3,p_4\in[0,1]$ are such that
$\Prob{E^{\mle}} \geq 1-p_1$,
$\Prob{E^{\bound}} \geq 1-p_4$,
and that for any given $ t >\tau$ and \emph{for any possible history $\cH_{t-1}$ before the start of round $t$,}
we have
\[
\Pr(E^{\anti}_t|\cH_{t-1}) \geq p_3,
\textnormal{ and }
\Pr(E^{\conc}_t|\cH_{t-1}) \geq 1-p_2.
\]
Then, the regret after $T$ rounds is bounded by
\begin{align*}
R(T) & \leq \cL \cdot \left(c_1 + \frac{c_2}{\sqrt \mu}\right) \left(1 + \frac{2}{p_3 - p_2} \right) \sqrt{c_3 T} + (p_1 + p_2 + p_4) \; T + \tau.
\end{align*}
\end{theorem}
Before proving this theorem, we show how it implies Theorem~\ref{thm:glb}.

\begin{proof}[Proof of Theorem~\ref{thm:glb}]

We first describe the initialization.

Define 
$\tau_0 \coloneqq \max \left\{
\frac{d \log(T/d) + 2 \log(T)}{\mu^2 \; \rho}, 1/\rho\right\}$.
First, we pull each $v_j$ for $\tau_0$ many times.
By then, the smallest eigenvalue of the Gram matrix $M_t$ becomes at least
$\rho \tau_0 = \max \left\{
\frac{d \log(T/d) + 2 \log(T)}{\mu^2 }, 1\right\}$,
so by the arguments in Lemma~8 from \citet{kveton2019randomized} and Theorem~1 from \citet{li2017provable}, $\norm{\hth - \sth} \leq 1$ in each subsequent round, with probability at least $1 - 1/T$.
In particular, by definition of $\mu$, this implies that in each subsequent round $t$ we have
$g'\left(\dop{X_t}{\thetahat_t}\right)\geq \mu$.
We pull each $v_i$ one more time.
After these pulls, we  have $H_t \succeq \mu \sum v_i v_i\transpose \succeq \mu \rho I \succ 0$.
Therefore, with $\tau = d + \max \left\{
\frac{d^2 \log(T/d) + 2d \log(T)}{\mu^2 \; \rho}, d/\rho\right\}$ initial
rounds, the event $E^{\bound}$ holds with probability at least $1-1/T = 1-p_4$.

Note that for $\randucb$, we have $\ftilde_t(x) = \langle \hth, x \rangle + Z_t \indnorm{x}{\invMt}$, so by definition, $\Prob{E^{\anti}_t} = \Prob{Z_t>\sqrt{\cL} c_1} \eqqcolon p_3$ and
$\Prob{\overline{E^{\conc}_t}} = \Prob{|Z_t|>c_2}\eqqcolon p_2$. 
Moreover, by Lemma~5 from \citet{kveton2019randomized} we have $\Prob{E^{\mle}}\geq 1-1/T=1-p_1$.
These bounds combined with Theorem~\ref{thm:genlinbanditsmainthm} give Theorem~\ref{thm:glb}.
\end{proof}

In the rest of this section we prove Theorem~\ref{thm:genlinbanditsmainthm}. Let use denote by $\E_t,\Pr_t$ the randomness injected by the algorithm in round $t$ (i.e., the randomness of $Z_t$), and denote by $\E_{\cH_{t-1}},\Pr_{\cH_{t-1}}$ the randomness in the history (up to round $t-1$). Let $\Delta_i = g(x_1, \sth) - g(x_i, \sth)$ denote the gap of the arm $i$ under the generalized linear model. The following lemma is an analogue of Lemma~2 from \citet{kveton2019randomized}.
\begin{lemma}
\label{lem:genlin-inst_regret}
For any round $t>\tau$ and any history $\cH_{t-1}$, we have
\[
 \E_t \; [\Delta_{i_t}\ind{E^{\mle}, E^{\bound}} | \cH_{t-1}] \leq p_2 + \cL \left(c_1 + \frac{c_2}{\sqrt\mu} \right) \left( 1+\frac{2}{p_3-p_2}\right) \E_t \; [\min\{1,\|X_t\|_{M_t^{-1}}\} | \cH_{t-1}].
\]
\end{lemma}
Before proving this lemma we show that it implies Theorem~\ref{thm:genlinbanditsmainthm}.
\begin{proof}[Proof of Theorem~\ref{thm:genlinbanditsmainthm}]
On the event $E^{\bound}$, for $t>\tau$ we have $M_t\succeq I$, hence by \eqref{19lemma3} and the Cauchy-Schwarz inequality, we have, deterministically,
\[\sum_{t=\tau+1}^T \min\{1,\|X_t\|_{M_t^{-1}}\} \leq \sqrt{c_3 T}.\]
Thus, since $\Delta_i\leq 1$ for all $i$, and using Lemma~\ref{lem:genlin-inst_regret} we have,
\begin{align*}
\textnormal{R(T)} & =  \sum_{t=  1}^T \E \; [\Delta_{i_t}] \\
& \leq \tau+ \sum_{t=\tau+1}^T \E \; [\Delta_{i_t}] \\
& \leq \tau+  T \; \P (\overline{E^{\mle}})
+  T \; \P (\overline{E^{\bound}})+ 
\sum_{t = \tau+1}^T \E \; [\Delta_{i_t}\ind{E^{\mle},E^{\bound}}]\\
&=  \tau+  T(p_1+p_4)+ 
\sum_{t=\tau+1}^{T} \E_{\cH_{t-1}} [ \E_t [\Delta_{i_t}\ind{E^{\mle}, E^{\bound}} | \cH_{t-1}] ]   \\
&\leq  T (p_1 + p_4) + \tau + \sum_{t=\tau+1}^{T}
\E_{\cH_{t-1}} \left[p_2 +  \cL \; \left( c_1 + \frac{c_2}{\sqrt\mu} \right) \left( 1+\frac{2}{p_3-p_2}\right) \E_t[\min\{1,\|X_t\|_{M_t^{-1}}\} | \cH_{t-1}] \right] \\
&=  T (p_1 + p_4) + \tau + \sum_{t=\tau+1}^{T}
\E \left[ p_2 + \cL \; \left( c_1 + \frac{c_2}{\sqrt\mu} \right) \left( 1+\frac{2}{p_3-p_2}\right) \min\{1,\|X_t\|_{M_t^{-1}}\} \right]\\
&= T (p_1 + p_2 + p_4) + \tau + 
\cL \; \left( c_1 + \frac{c_2}{\sqrt\mu} \right) \left( 1+\frac{2}{p_3-p_2}\right) \E \sum_{t=\tau+1}^{T} \left[\min\{1,\|X_t\|_{M_t^{-1}}\} \right]\\
& \leq T (p_1 + p_2 + p_4) + \tau + 
\cL \; \left( c_1 + \frac{c_2}{\sqrt\mu} \right) \left( 1+\frac{2}{p_3-p_2}\right) \sqrt{c_3 T},
\end{align*}
completing the proof of the theorem.
\end{proof}

We finally prove Lemma~\ref{lem:genlin-inst_regret}.
\begin{proof}[Proof of Lemma~\ref{lem:genlin-inst_regret}]
Fix a round $t$ and an arbitrary history $\cH_{t-1}$ satisfying $E^{\mle}$ and $E^{\bound}$, and omit the conditioning on $\cH_{t-1}$ henceforth. 
On the event $E^{\bound}$ we have $M_t \succeq I$, and since $\|X_t\|\leq1$ we have
$\|X_t\|_{M_t^{-1}}\leq1$ deterministically,
so we need only show that
\[
 \E \; [\Delta_{i_t}\ind{E^{\mle}, E^{\bound}}] \leq p_2 + \cL \left(c_1 + \frac{c_2}{\sqrt\mu} \right) \left( 1+\frac{2}{p_3-p_2}\right) \E\; [\|X_t\|_{M_t^{-1}}].
\]
Let us define 
\begin{align*}
h(x)&\coloneqq c_1 \indnorm{x}{\invMt} + c_2 \indnorm{x}{\invHt},\\ 
\widetilde{\Delta}_i & \coloneqq \dop{x_1-x_i}{\thetastar},\\
S_t & \coloneqq \{i\in \cA : h(x_i) < \widetilde\Delta_i\}
\textnormal{ and }
\overline{S_t} = \cA \setminus S_t.
\end{align*}
Observe that $S_t$ is deterministic (since we have fixed the history) and that $1 \notin S_t$.
The arms in $S_t$ are sufficiently sampled, and the rest of the arms are undersampled.
Also, let 
\[
j_t \coloneqq \argmin_{i\notin S_t} h(x_i)
\]
be the least uncertain undersampled arm, which is deterministic since we have fixed the history.  We may then write
\begin{align}
\E [\Delta_{i_t}] &=
\E [\Delta_{i_t} \ind{E^{\conc}_t}]
+ \E [\Delta_{i_t} \ind{\overline{E^{\conc}_t}}] \notag\\
&\leq \E [\Delta_{i_t} \ind{E^{\conc}_t}]
+ \Prob{\overline{E^{\conc}_t}} \notag \\
&\leq \E [\Delta_{i_t} \ind{E^{\conc}_t}] + p_2 \notag \\
& = \E [g(\dop{x_1}{ \sth}) - g(\dop{x_{i_t}}{ \sth}) \ind{E^{\conc}_t}] + p_2 \notag \\
& \leq \cL \; \E [(\dop{x_1}{ \sth} - \dop{x_{i_t}}{ \sth}) \ind{E^{\conc}_t}] + p_2 \notag \\ \implies
\E [\Delta_{i_t}] & \leq \cL \; \E [\widetilde{\Delta}_{i_t} \ind{E^{\conc}_t}] + p_2. \label{deltapartitiongen}
\end{align}

Recall that $E^{\conc}_t$ is the event that for all arms $i$,
$\left|\ftilde_t(x_i) - \dop{x_i}{\thetahat_t}\right| \leq c_2 \|x_i\|_{\invHt}$,
and $E^{\mle}$ is the event that for all arms $i$,
$\left| \dop{x_i}{\thetahat_t-\thetastar}\right|\leq c_1 \|x_i\|_{\invMt}$.
Hence, on event $E^{\conc}_t\cap E^{\mle}$ we have
\[
\ftilde_t(x_{i_t})
\leq c_2 \|x_{i_t}\|_{\invHt} + 
\dop{x_{i_t}}{\thetahat_t}
\leq c_2 \|x_{i_t}\|_{\invHt} + 
c_1 \|x_{i_t}\|_{\invMt} +
\dop{x_{i_t}}{\theta^*}
\implies \ftilde_t(x_{i_t}) \leq h(x_{i_t}) + \dop{x_{i_t}}{\theta^*},
\]
and, similarly,
\[
\ftilde_t(x_{j_t}) \geq 
\dop{x_{j_t}}{\theta^*} - h(x_{j_t}),
\]
which, since $\ftilde_t(x_{i_t})\geq
\ftilde_t(x_{j_t})$, gives
\begin{align*}
\widetilde{\Delta}_{i_t}&=\widetilde{\Delta}_{j_t}+
\dop{x_{j_t}-x_{i_t}}{\thetastar} \leq h(x_{j_t}) + h(x_{i_t}) + h(x_{j_t})=h(x_{i_t}) + 2 \; h(x_{j_t})
\end{align*}
deterministically.
Plugging into \eqref{deltapartitiongen}, we obtain
\begin{equation}\label{expecteddeltagen}
\E[\Delta_{i_t}] \leq \cL \; \left[\E \; h(x_{i_t}) + 2  h(x_{j_t}) \right] + p_2.
\end{equation}
The next step is to bound $h(x_{j_t})$ from above.
Observe that
\[
\E h(x_{i_t})
\geq
\Ex{ h(x_{i_t}) | i_t \in \overline{S_t}}\Prob{i_t \in \overline{S_t}}
\geq
h(x_{j_t}) \Prob{i_t \in \overline{S_t}},
\]
where the last inequality is by the definition of $j_t$.
Rearranging gives
\[
h(x_{j_t})
\leq
\E \; h(x_{i_t})
\big/\Prob{i_t \in \overline{S_t}}.
\]
Next, we bound $\Prob{i_t \in \overline{S_t}}$ from below.
By definition of $i_t$ and since $1\in\overline{S_t}$, we have
\[
\Prob{i_t \in \overline{S_t}}
\geq
\Prob{\ftilde_t(x_1) > \max_{j\in S_t}\ftilde_t(x_j)}
\geq
\Prob{\ftilde_t(x_1) > \max_{j\in S_t}\ftilde_t(x_j) \textnormal{ and } E^{\conc} \textnormal{ holds}}.
\]
If $E^{\conc}$ holds, then for any $j\in S_t$ we have
\[
\ftilde_t(x_j) \leq \dop{x_j}{\thetastar} + h(x_j)
<
\dop{x_j}{\thetastar} + \widetilde\Delta_j
= \dop{x_1}{\thetastar},
\]
whence,
\begin{align*}
\Prob{\ftilde_t(x_1) > \max_{j\in S_t}\ftilde_t(x_j) \textnormal{ and } E^{\conc} \textnormal{ holds}}
& \geq
\Prob{\ftilde_t(x_1) >  \dop{x_1}{\thetastar} \textnormal{ and } E^{\conc} \textnormal{ holds}}\\
&\geq
\Prob{\ftilde_t(x_1) >  \dop{x_1}{\thetastar}} - \Prob{\overline{E^{\conc}}}.
\end{align*}
Since $E^{\mle}$ holds,
\[
\Prob{\ftilde_t(x_1) >  \dop{x_1}{\thetastar}} \geq \Prob{\ftilde_t(x_1) - \dop{x_1}{\hth} > c_1 \indnorm{x_1}{\invMt}}.
\]
By Lemma~\ref{lem:MtHtrelation} below we have 
\(\indnorm{x_1}{\invMt} \leq \sqrt{\cL} \; \indnorm{x_1}{\invHt} \), implying
\[
 \Prob{\ftilde_t(x_1) - \dop{x_1}{\hth} > c_1 \indnorm{x_1}{\invMt}} \geq \Prob{\ftilde_t(x_1) - \dop{x_1}{\hth} > c_1 \sqrt{\cL} \; \indnorm{x_1}{\invHt}}=\Prob{E^{\anti}}\geq p_3,
\]
whence,
\[
 \Prob{\ftilde_t(x_1) >  \dop{x_1}{\thetastar}}- \Prob{\overline{E^{\conc}}}
\geq p_3-p_2.
\]
Hence, we find
$\Prob{i_t \in \overline{S_t}}\geq p_3-p_2$, and plugging this back into
\eqref{expecteddeltagen} gives
\[
\E [\Delta_{i_t}] \leq \cL \; \left[\E h(x_{i_t}) + \frac{2 \; \E \; h(x_{i_t})}{p_3 - p_2} \right] + p_2 = \cL \left[1 + \frac{2}{p_3 - p_2} \right] \; \E h(x_{i_t}) + p_2.
\]
We now bound the quantity $h(x_{i_t})$. By Lemma~\ref{lem:MtHtrelation} below we have 
\[
\indnorm{X_t}{\invHt} \leq \frac{\indnorm{X_t}{\invMt}}{\sqrt\mu}, \]
thus,
\[
 h(X_t) = c_1 \indnorm{X_t}{\invMt} + c_2 \indnorm{X_t}{\invHt} \leq \left(c_1 + \frac{c_2}{\sqrt\mu} \right) \indnorm{X_t}{\invMt}.
\]
Putting everything together, we find
\[
\E [\Delta_{i_t}] \leq \cL \; \left[1 + \frac{2}{p_3 - p_2} \right] \; \left(c_1 + \frac{c_2}{\sqrt\mu} \right) \E \indnorm{X_t}{\invMt} + p_2,
\]
completing the proof of Lemma~\ref{lem:genlin-inst_regret}.
\end{proof}

\begin{lemma}\label{lem:MtHtrelation}
Let $V_1,\dots,V_t$ be positive semi-definite matrices,
and let $\mu \leq b_1,\dots,b_t \leq \cL$ be  real numbers
such that both
$A \coloneqq \sum_{s=1}^{t} V_t$
and
$B \coloneqq \sum_{s=1}^{t} b_t V_t$
are positive definite.
Then, for any vector $x$ we have
\[
\sqrt \mu \|x\|_{B^{-1}}
\leq
\|x\|_{A^{-1}}
\leq
\sqrt{\cL} \|x\|_{B^{-1}}.
\]
\end{lemma}

\begin{proof}
Note that 
$\cL A = \sum_{s=1}^{t} \cL V_t
\succeq \sum_{s=1}^{t} b_t V_t = B$.
Invertibility of the PSD ordering implies
$A^{-1}/\cL \preceq B^{-1}$,
whence for any vector $x$ we have
$x\transpose A^{-1} x \leq \cL x\transpose B^{-1} x$, which gives the  inequality on the right.
The left inequality is proven symmetrically, by choosing $V'_t\coloneqq b_t V_t$ and $b'_t \coloneqq 1/b_t$.
\end{proof}
\section{Additional experiments}
\label{app:expes}

All algorithms are compared based on their cumulative empirical regret, defined as
\begin{align*}
    R(T) := \sum_{t=1}^T \big( Y_t^\star - Y_t \big),
\end{align*}
where $Y_t^\star$ denotes the reward of the optimal arm in round $t$.

In all experiments, algorithms are run over $T=20,000$ rounds on 50 randomly generated instances (the generated instances are the same for all algorithms). 

\subsection{Ablation study}
\label{app:expes:ablation}

We first investigate the impact of the $\randucb$ design choices and parameters on the MAB settings with, unless specified, $L=0$, $U=2\sqrt{\ln(T)}$, and $M=20$.

\begin{figure}[h]
    \centering
    \includegraphics[width = 0.95\textwidth]{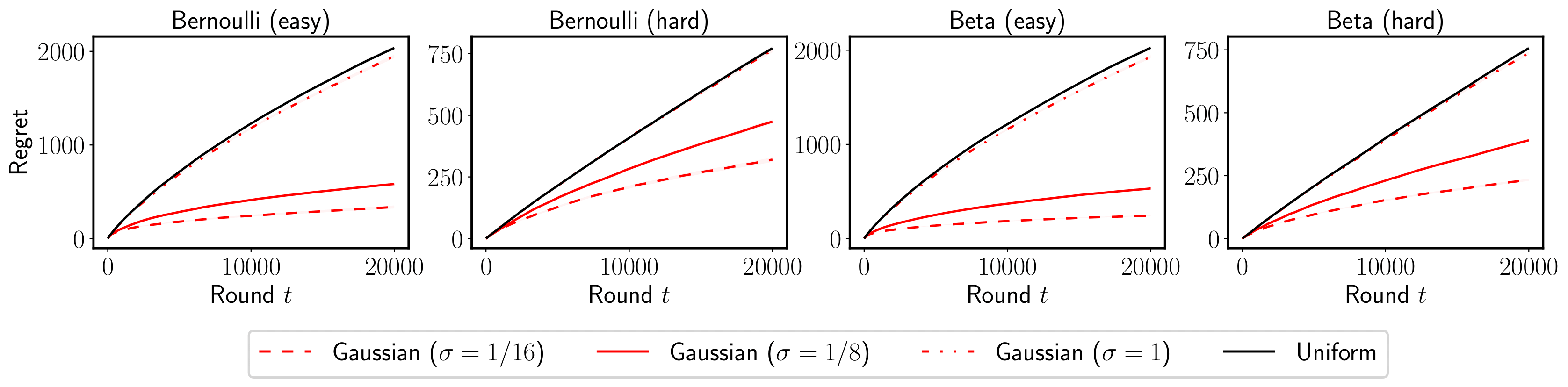}
    \caption{Cumulative empirical regret with different sampling distributions.}
    \label{fig:results:ablation:sampling_distribution}
\end{figure}

Recall that $\randucb$ is characterized by the choice of sampling distribution (\cref{sec:randucb:sampling_dist}). We compare the performance of $\randucb$ using uniform and Gaussian sampling ($\epsilon=10^{-7}$, $\sigma\in\{1/16, 1/8, 1\}$) distributions. Figure~\ref{fig:results:ablation:sampling_distribution} shows that increasing $\sigma$ brings us closer to the uniform distribution.

\begin{figure}[h]
    \centering
    \includegraphics[width = 0.95\textwidth]{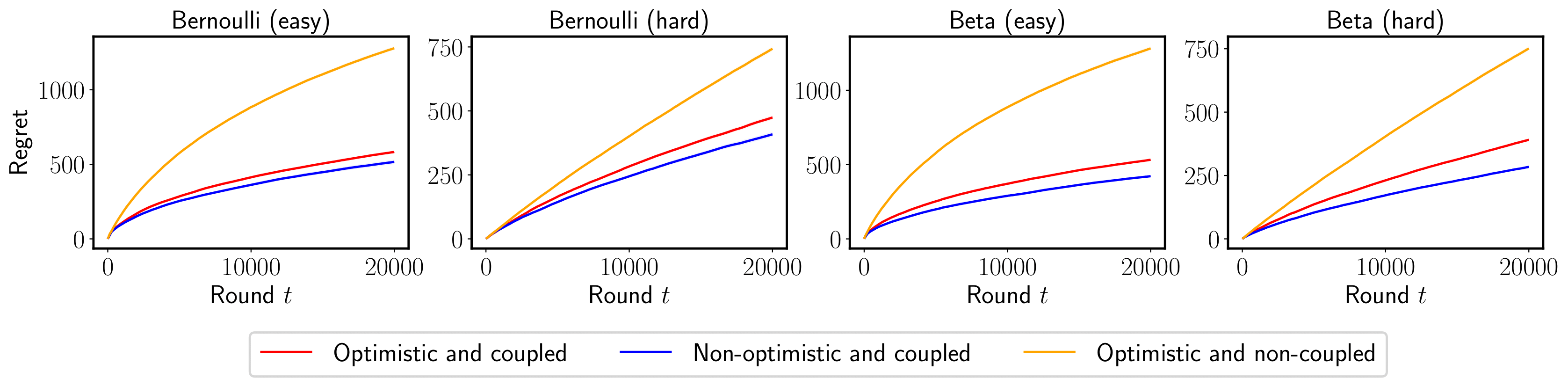}
\caption{Cumulative empirical regret with different configurations of optimism and arms coupling.}
\label{fig:results:ablation:optimism_coupling}
\end{figure}

We then compare the default (optimistic, with coupled arms) $\randucb$ with non-optimistic and uncoupled variants. All use Gaussian sampling ($\epsilon=10^{-7}$, $\sigma=1/8$). Non-optimistic considers $L=-2\sqrt{\ln(T)}$ and $M=40$. Figure~\ref{fig:results:ablation:optimism_coupling} shows that coupling the arms is more determinant in the performance of $\randucb$ compared with optimism. This is not surprising as the same happens for TS (\cref{app:expes:ots}).

\begin{figure}[h]
    \centering
    \includegraphics[width = 0.95\textwidth]{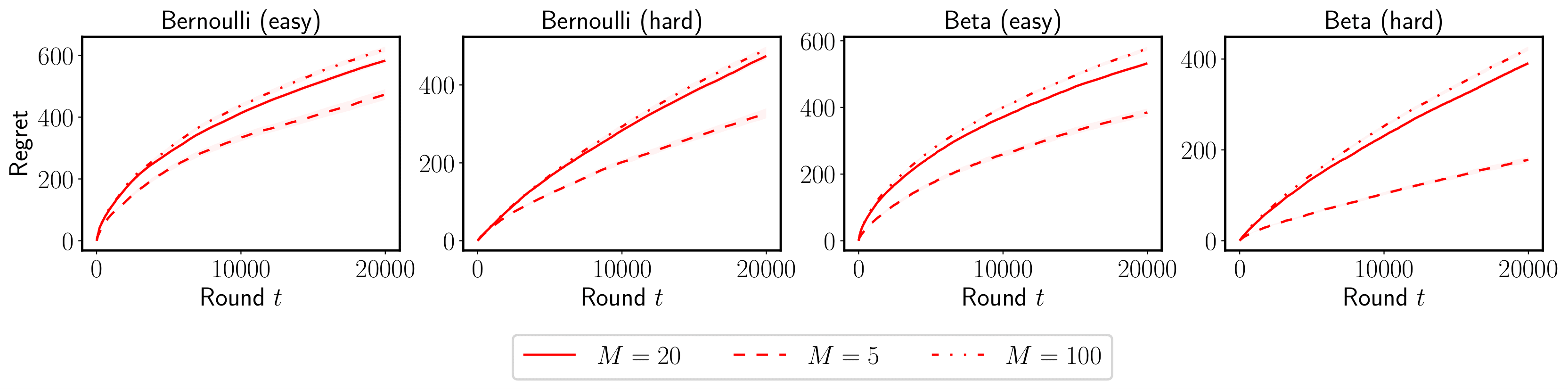}
\caption{Cumulative empirical regret for $\randucb$ with different sampling distribution support size $M$.}
\label{fig:results:ablation:nbins}
\end{figure}

We  also  evaluate  the  impact  of  the  support size of the sampling distribution $M$. We compare $\randucb$ using Gaussian sampling distribution $(\epsilon=10^{-7},\sigma= 1/8)$, $M=20$,  optimistic, and coupled arms, against alternatives with $M=5$ and $M=100$. Figure~\ref{fig:results:ablation:nbins} shows that $\randucb$ is robust to the discretization induced by the support size.

\subsection{Optimistic Thompson Sampling}
\label{app:expes:ots}

\begin{figure}[h]
    \centering
    \includegraphics[width = 0.95\textwidth]{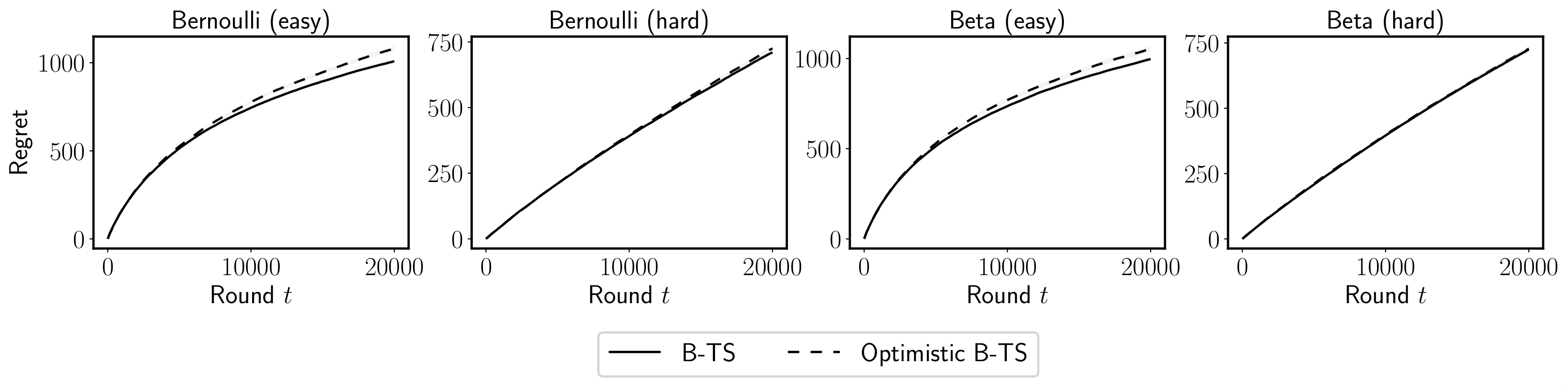}
\caption{Cumulative empirical regret for optimistic TS versus TS.}
\label{fig:results:optimistic_ts}
\end{figure}

In the last experiment, we empirically show that Optimistic TS is almost equivalent to TS. To this end, we compare both variants on the MAB setting. Figure~\ref{fig:results:optimistic_ts} confirms that their performance is similar.

\subsection{Gaussian MAB}
\label{app:expes:gaussian}
We use the same number of arms $K = 100$. The rewards are generated from a $N(\mu, 0.1)$ distribution. For generating the mean rewards, we consider both the easy and hard settings as before. 
\begin{figure}[h]
    \centering
    \includegraphics[width = 0.95\textwidth]{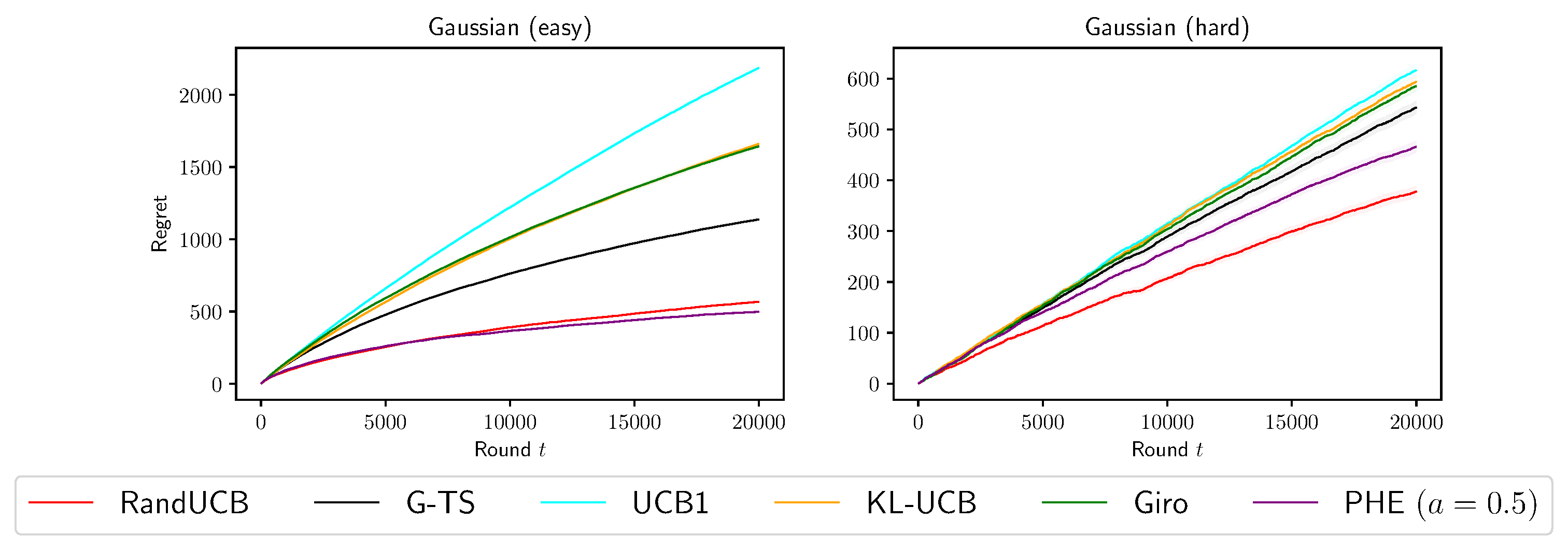}
\caption{Cumulative empirical regret for Gaussian MAB.}
\label{fig:results:gaussian-mab}
\end{figure}
Both Gaussian TS and UCB1 are theoretically optimal in this setting. From ~\cref{fig:results:gaussian-mab}, we observe that similar to the other settings, the performance of RandUCB (with the same hyperparameter settings) is better than both TS and UCB1. 

\end{document}